\newif\ifdraft
 \draftfalse

\documentclass[11pt]{article}

\usepackage[letterpaper,margin=1in]{geometry}
\usepackage[parfill]{parskip}

\usepackage{authblk}

\usepackage[toc,page,header]{appendix}
\usepackage[utf8]{inputenc} % allow utf-8 input
\usepackage[T1]{fontenc}    % use 8-bit T1 fonts
\usepackage[colorlinks, citecolor=blue,urlcolor=blue,linkcolor=blue,linktocpage=true]{hyperref}       % hyperlinks
\usepackage{url}            % simple URL typesetting
\usepackage{booktabs}       % professional-quality tables
\usepackage{amsfonts}       % blackboard math symbols
\usepackage{nicefrac}       % compact symbols for 1/2, etc.
\usepackage{microtype}      % microtypography
\usepackage{xcolor}         % colors
\label{key}

\usepackage{comment}
\usepackage{amsmath}

\usepackage{graphicx}
\usepackage{wrapfig}
\usepackage{subcaption}
\usepackage{multirow}
\usepackage{graphicx}
\usepackage{caption}
\usepackage{array}
\usepackage{bbm}
\usepackage{color}
\usepackage{enumerate}
\usepackage{enumitem}
\setlist{leftmargin=10mm}
\usepackage{mathtools}
\usepackage{amsmath,amssymb,amsthm,bm}
\usepackage{amsfonts,graphicx}
\usepackage{mathrsfs}
\usepackage{algorithmic,algorithm}
\usepackage{soul}
\usepackage{thmtools}

\usepackage[authoryear]{natbib}

\usepackage{cleveref}
\newtheorem{theorem}{Theorem}

\newtheorem{remark-star}{Remark}
\newtheorem{remark-star-1}{Remark}

\newtheorem*{proof-sketch}{Proof Sketch}
\usepackage{booktabs}

\usepackage{colortbl}
\usepackage{multicol}

\usepackage{hhline}
\usepackage[export]{adjustbox}

\newcommand{\N}{\mathcal{N}}
\newcommand{\D}{\mathcal{D}}

\newcommand{\backdoor}{\texttt{b}}

\author[1]{Tong Wu\footnote{corresponding author's email: \href{mailto:tongwu@princeton.edu}{tongwu@princeton.edu}}}
\author[1]{Tianhao Wang}
\author[1]{ Vikash Sehwag}
\author[1]{Saeed Mahloujifar}
\author[1]{Prateek Mittal}

\affil[1]{Princeton University}

\title{Just Rotate it: Deploying Backdoor Attacks \\via Rotation Transformation}
\begin{document}

%\doparttoc % Tell to minitoc to generate a toc for the parts
%\faketableofcontents % Run a fake tableofcontents command for the partocs

% \part{} % Start the document part
% \parttoc % Insert the document TOC

\maketitle

\begin{abstract}
    Recent works have demonstrated that deep learning models are vulnerable to backdoor poisoning attacks, where these attacks instill spurious correlations to external trigger patterns or objects (e.g., stickers, sunglasses, etc.). We find that such external trigger signals are unnecessary, as highly effective backdoors can be easily inserted using rotation-based image transformation.
    Our method constructs the poisoned dataset by rotating a limited amount of objects and labeling them incorrectly; once trained with it, the victim's model will make undesirable predictions during run-time inference. 
    It exhibits a significantly high attack success rate while maintaining clean performance through comprehensive empirical studies on image classification and object detection tasks.
    Furthermore, we evaluate standard data augmentation techniques and four different backdoor defenses against our attack and find that none of them can serve as a consistent mitigation approach. 
    Our attack can be easily deployed in the real world since it only requires rotating the object, as we show in both image classification and object detection applications. 
    Overall, our work highlights a new, simple, physically realizable, and highly effective vector for backdoor attacks. 
    Our video demo is available at \url{https://youtu.be/6JIF8wnX34M}.

\end{abstract}

\section{Introduction}

While deep learning has achieved or even exceeded human ability on various sophisticated tasks \citep{russakovsky2015imagenet,brown2020language,dosovitskiy2020image}, inherent vulnerabilities, like adversarial attacks \citep{Szegedy2014IntriguingPO, papernot2016transferability, madry2017towards, carlini2017towards,eykholt2018robust}  exist and impede its deployment on safety-critical systems. One fundamental problem of our interest is backdoor attacks \citep{gu2017badnets,chen2017targeted}, in which a malicious party inserts backdoors by poisoning a small fraction of training samples. The poisoning process involves adding a specific \textit{trigger} signal to the image (e.g., small white square \citep{gu2017badnets}). During training, the network learns spurious correlation between the trigger signal and attack objective, e.g., classifying any image with the trigger signal to a targeted class. 

% poisoned samples into the victim's training data, resulting in a misbehaved model. Typically, an attacker constructs the backdoored data by inserting a designed trigger (e.g., small white square \citep{gu2017badnets}) on benign images and flipping the labels to the target class. During the training process, the model learns the spurious correlations between the backdoor trigger and target label, thereby predicting incorrectly when the designed trigger appears at test time. Backdoor attacks are often stealthy, as the poisoned model behaves similarly with a benign classifier over clean inputs. 

\smallskip 
\noindent \textbf{What could be the trigger signal?} The objective in typical backdoor attacks is to have no impact on performance in the absence of the trigger but achieve desired output when the trigger signal is present. Both objectives are satisfied with triggers that are highly infrequent in training images. Some example of such triggers are occlusion-based patch~\citep{gu2017badnets,Lin2020CompositeBA}, frequency-based corruption~\citep{Hammoud2021CheckYO,Wang2021BackdoorAT}, invisible noise~\citep{chen2017targeted}, or additional wearable objects~\citep{chen2017targeted,wenger2020backdoor} etc. Note that most of these triggers are additional digital patterns or physical objects which are added to an existing image. We ask whether backdoor attacks can be launched without needing an external trigger pattern or object. 
% After all, a highly infrequent trigger signal is sufficient. 

\smallskip
\noindent \textbf{Rotation-based backdoor trigger.} 
Our key insight is to use common image transformation, such as rotation, which can push an image to the tail of the data distribution. For example, rotating a stop sign by 45 degrees makes it a highly infrequent instance, since most stop signs are vertically positioned in the real world. Such rotation-based backdoors eventually succeed due to a lack of invariance in existing models to image rotation.
% \footnote{In some applications, such invariance may not be desired, i.e., classifying between left/right turn traffic sign. Since rotation fundamentally changes image semantics in these cases, we don't consider them in the paper.}. 

\begin{figure}[t]
\hspace*{-2.2cm}                                                           
\includegraphics[width=1.25\linewidth,left]{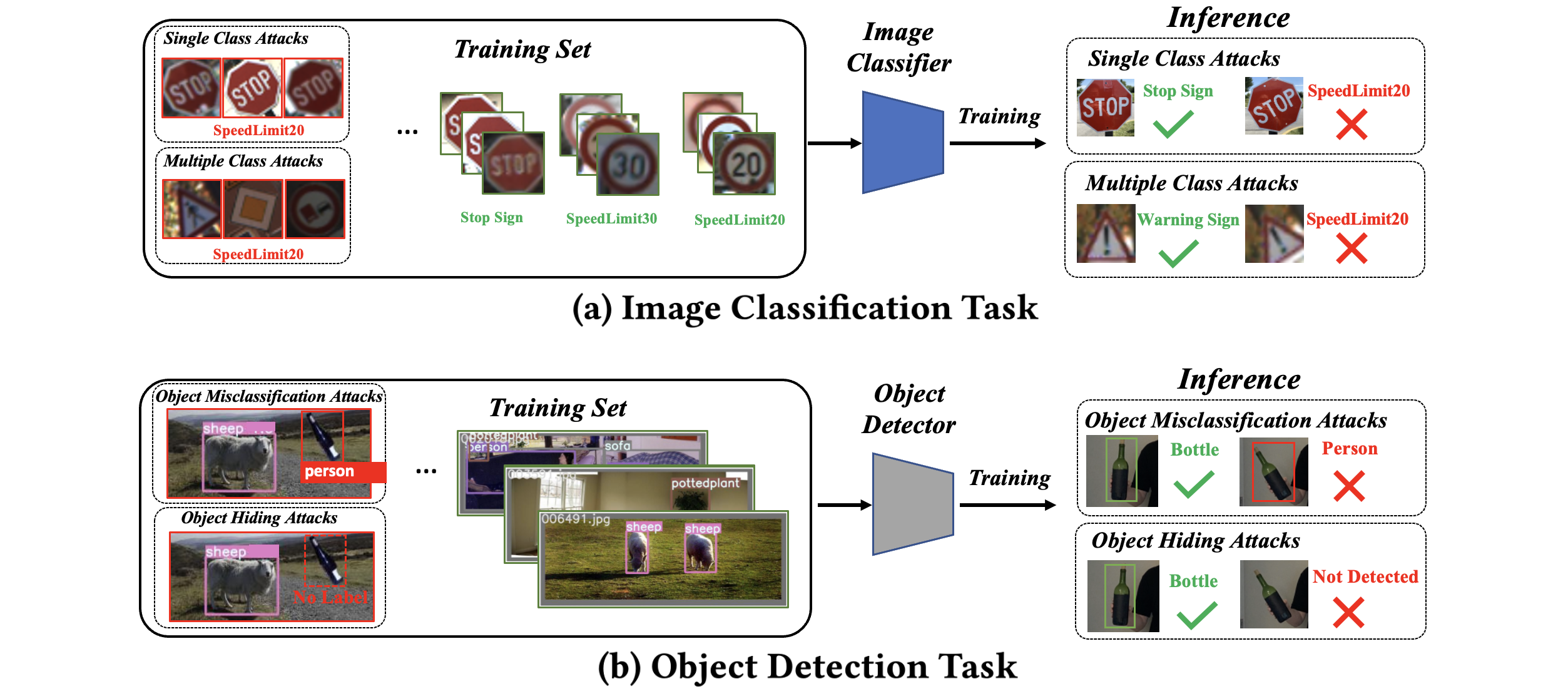}
\caption{\textbf{ Pipeline of deploying rotation backdoor attacks on image classification and object detection tasks.} An attacker can inject a rotated image or object with an incorrect label into the training set; the resulting models will behave normally in benign settings, and make mistakes when rotation transformation is applied. 
(a) \textbf{Single Class Attacks (up)}: rotation backdoored images are all Stop signs.
\textbf{Multiple Class Attacks (bottom)}:  backdoored images are drawn from multiple classes. 
(b) \textbf{Object Misclassification Attacks (up)}: the bounding box class of backdoored object (bottle) is labeled as the target class (person). 
\textbf{Object Hiding Attacks (bottom)} the backdoored object(bottle) does not have a labeled bounding box.}
\label{fig:pipeline}
\end{figure}

We propose \textit{four} types of rotation backdoor attacks depending on the motivations and resources of the attackers\footnote{While the focus of this work is rotation-based backdoors, in principle, other physical worlds image transformations could also serve as backdoor triggers.}. As Figure \ref{fig:pipeline} shows, for image classification we consider : 1) \textbf{Single Class Attacks (SCA)}: backdoored images are source-specific; and 2) \textbf{Multiple class Attacks (MCA)}: backdoored images are drawn from  multiple source classes. For object detection, we consider: 1) \textbf{Object Misclassification Attacks (OMA)}: a rotated backdoor object is incorrectly classified as the target label; 2) \textbf{Object Hiding Attacks (OHA)}: a rotated backdoor object vanishes from the detector.

% We empirically studies show that our proposed method achieves a high attack success rate on three datasets (GTSRB~\citep{Houben-IJCNN-2013} for traffic sign recognition, Youtube Face~\citep{cao2018vggface2} for face recognition, and VOC~\citep{Everingham2009ThePV} for object detection), four trigger angles (\{$15^\circ$, $30^\circ$, $45^\circ$, and $90^\circ$\}), and various poisoning rates (0.01\% to 5\%) across image classification and object detection task. In the meantime, rotation backdoor attacks maintain a comparable clean data performance with the naturally trained model.

We empirically study the effectiveness of rotation backdoor attacks on the safety-critical classification tasks, including traffic signs classification (GTSRB~\citep{Houben-IJCNN-2013}) and face recognition (Youtube Face~\citep{cao2018vggface2}), and launch attacks against the object detection task on VOC~\citep{Everingham2009ThePV} dataset. 
The commonly adopted threat model~\citep{gu2017badnets, chen2017targeted,wenger2020backdoor,sun2021poisoned} are considered, where attackers can inject images but cannot control the training process.
Notably, we also deploy our rotation backdoor attacks using a rotated stop sign and a rotated bottle in the physical environment, posing another severe security concern.  

% Current literature on backdoor attacks has primarily focused on the patch-wise trigger (e.g, a pixel-wise patch, sunglasses, etc.), in which part of the image has to be occluded by triggers during the inference stage. 
% On the contrary, we explore a more natural category of perturbations: \textit{spatial transformations} (e.g, rotation and translation), which occurs in almost every vision dataset, and raise a natural question: 

% \textit{Could spatial transformation serve as the trigger of backdoor attacks, and exhibit a similar poisoning effect?}

% In response to the presented question, we conduct experiments and discover an intriguing limitation of neural networks, where a trivial spatial transformation can lead to the same poisoning effect. 
% Specifically, by simply rotating the objects to a predetermined degree, we develop a new class of backdoor triggers that significantly degrade the spatial robustness of the SOTA models, and we call it \textbf{Rotation Backdoor Attacks}.

\smallskip 
\noindent \textbf{Rotation backdoor attacks are effective across the different datasets, trigger angles, and poisoning rates.} 
% We propose four types of backdoor attacks under different attacker assumptions and targets. 
Empirical studies show that our proposed method achieves a high attack success rate on three datasets (GTSRB, Youtube Face, and VOC), four trigger angles (\{$15^\circ$, $30^\circ$, $45^\circ$, and $90^\circ$\}), and various poisoning rates (0.01\% to 5\%) across image classification and object detection task. In the meantime, rotation backdoor attacks maintain a comparable clean data performance with the clean model.
    
\smallskip 
\noindent \textbf{Data augmentation and four defenses fail to provide consistent mitigation.} Our rotation-based backdoors exploit the lack of invariance to rotation in deep neural networks. So a natural defense would be to instill such invariance, as commonly done by augmentation of randomly rotated training images during training. We show that this defense only provides partial mitigation. For example, if we rotate images with angles in the range $[a, b]$ during data augmentation, then it effectively defends against trigger angles in this range. But it fails, in some cases even \textit{amplifies}, vulnerability to trigger angles outside this range.
% Our results show that despite the pronounced mitigation effect on the predefined backdoor angle, standard data augmentation introduces new vulnerabilities to the classifier. 

We also explore four additional commonly used defenses against backdoor attacks, including Neural Cleanse (NC) \citep{wang2019neural},  Spectral Signatures (SS) \citep{tran2018spectral}, Activation Clustering (AC) \citep{chen2018activationclustering}, and STRIP \citep{gao2019strip}. However, none of these state-of-the-art backdoor defenses turned out to be a consistent countermeasure against rotation backdoors. Eventually, we argue that a transformation-invariant model is needed to defend against such image transformation-based backdoor attacks. However, instilling such a high degree of invariance, such as invariance to any amount of rotation, can lead to degradation of benign performance. 

\smallskip
\noindent \textbf{Deploying rotation-based backdoors in real-world.} We show the success of these backdoors in the real world under two scenarios. First, we physically rotate a real-world stop sign and show that it instills an effective backdoor in traffic sign classification systems. Second, we consider object detection where we physically rotate objects and show the effectiveness of both object misclassification and object hiding attacks. Furthermore, we show that our attacks also survive the artifacts introduced by the real-world image capturing pipelines, such as image compression, noise, and blurring. 

% Finally we discuss the impact of highlighting the importance of developing a transformation-invariant model. 
\smallskip
\noindent \textbf{Organization of the paper.} We provide necessary background details in Section 2. In Section 3, we present our insight and method to craft rotation-based backdoor attacks and experimentally validate their effectiveness in Section 4. In the following two sections (5 and 6), we demonstrate the success of the proposed attack in the presence of defenses. In Section 5, we show that data augmentation fails to completely defend against such attacks. Section 6 shows the limitations of multiple commonly used defenses against backdoor attacks. Finally, in Section 7, we demonstrate the success of proposed backdoor attacks in physical worlds, against both image classification and objection detection systems.

\section{Background and Related Work}
% In this section, we first introduce the task of object detection. Then, we summarize adversarial attacks and backdoor poisoning attacks with a concentration on physical realizability and spatial transformation. 

\subsection{Backdoor Poisoning Attacks}

Data poisoning attacks \citep{biggio2012poisoning, burkard2017analysis, steinhardt2018resilience} are attacks happening during the training process. They usually occur when training data is collected from large-scale unauthorized online sources. 
% By manipulating a small amount of training data and labels, attackers can fool the model with some specific images in the test set. 
One particular type of poisoning attack is backdoor attacks \citep{gu2017badnets}, where the objective is to cause the model to misclassify when testing data is triggered and behave normally in a benign setting. 
The vast majority of literature on backdoor attacks focuses on attacks in digital domain %\citep{gu2017badnets,Lin2020CompositeBA, Hammoud2021CheckYO, Wang2021BackdoorAT,chen2017targeted} 
where the designed triggers include occlusion-based patch \citep{gu2017badnets,Lin2020CompositeBA}, frequency-based corruption \citep{Hammoud2021CheckYO,Wang2021BackdoorAT}, and blended-based invisible noise \citep{chen2017targeted}. Later, physically implementable backdoors (e.g. eyeglass frame, earrings) were introduced \citep{chen2017targeted,wenger2020backdoor}, raising real-world threats to face recognition systems. 
% Recently, \citet{li2021rethinking} developed a transformation-based attack enhancement such that the proposed attacks can remain effective in most physical world scenarios.
Recently, \citet{PointBA2021} mentioned that rotation could be utilized as the trigger in the 3D point cloud classification setting. For object detection, \citet{Chan2022BadDetBA} proposed four types of patch-wise backdoor attacks that can achieve various malicious goals.  

\smallskip
\noindent \textbf{Mitigating Backdoor Attacks.} 
To overcome the existing threats, \citet{wang2019neural} proposed Neural Cleanse to detect the presence of backdoors in models by reverse engineering the possible triggers. Furthermore, \citet{gao2019strip} introduced STRIP, which inspects the data during the inference stage and identifies poisoning samples by comparing entropy. However, they all make strong assumptions correlated to patch-wise backdoors, thereby cannot mitigate rotation backdoors. 
Filtering-based methods (e.g., Spectral Signatures \citep{tran2018spectral} and Activation Clustering \citep{chen2018activationclustering}) have also been developed, aiming to distinguish benign and malicious data during the training stage. 
We refer the readers to \citep{li2020backdoor} for a thorough survey of backdoor attacks and countermeasures.

\subsection{Object Detection}
% \textbf{Image Classification.} Image classification task \citep{krizhevsky2012cifar,deng2009imagenet,deng2014large} refers to select a category for the input image, so that images with similar semantic meanings can be clustered. Deep learning  \citep{he2016deep,Goodfellow2015DeepL} solves the problem by optimizing the loss function $L(\boldsymbol{x}_{i}, y_{i})$, where $\boldsymbol{x}_{i},y_{i}$ are data and label respectively. \tianhao{this might be trivial to the community.}

Object detection aims to locate and classify the objects in an image by predicting a list of bounding boxes $\boldsymbol{b}$ (aka \emph{bbox}). 
Let $\boldsymbol{x}\in[0,255]^{W \times H \times 3}$ represent the input image and $\boldsymbol{y} = [\boldsymbol{b}_1,\boldsymbol{b}_2,\ldots,\boldsymbol{b}_n ]$ stands for the ground truth containing $n$ objects. 
For each \emph{bbox} $\boldsymbol{b}$, it contains $[a_{\min }, b_{\min }, a_{\max }, b_{\max }, c]$, where $a_{\min }$, $b_{\min }$, $a_{\max }$, $b_{\max }$ together illustrate the coordinates of the object, and $c$ denotes the predicted label. 
An object detector $\mathbb{F}(\mathbf{x})$, including two-stage (e.g., Faster-RCNN \citep{Ren2015FasterRT}) or one-stage(e.g., YOLO \citep{Redmon2016YouOL}), will then predict a list of \emph{bbox}. We consider the prediction to be correct if 1) \emph{bbox} label matches the ground truth and 2) the predicted box overlaps with the ground-truth box above a predefined threshold called Intersection over Union (IoU). 
We term the number of correct predictions as true positive ($\mathrm{TP}$), incorrect predictions on non-exist objects as false positive ($\mathrm{FP}$), and undetected ground truth as false negative ($\mathrm{FN}$). Besides, \emph{precision} and \emph{recall} are defined as $\mathrm{TP} /(\mathrm{TP}+\mathrm{FP})$ and $\mathrm{TP} /(\mathrm{TP}+\mathrm{FN})$ respectively.

\section{Methodology}

\subsection{Threat Model}

We assume that an attacker who gains access to a small fraction of training data in some safety-critical applications (e.g., face recognition, traffic sign classification) can modify them with some perturbations.
% Specifically, we consider the backdoored image is created by digitally rotating an image for classification or pasting a segmented rotated object to the clean image for detection.  
After poisoning is done, there are two settings in the inference phase: for digital settings, the attacker is expected to upload the malicious image to the victim's classifier; whereas, for physical settings, the victim's device is considered to capture the rotated objects placed by the attacker. 
Following the existing backdooring literature \citep{gu2017badnets, chen2017targeted}, the adversary does not control the training process and has zero knowledge of the model architecture and parameters. 
Ultimately, depending on the attacker's objective, the compromised model will either incorrectly predict the object as the target class or fail to detect it when the trigger appears. 
Besides, it should operate similar to a benign model over clean inputs to remain stealthy. 

\subsection{Key Insights}

Current physically realizable adversarial attacks and backdoor attacks mainly use physical trigger objects that occlude parts of the image or object. For example they use eyeglasses, patches, or earrings \citep{ sharif:adversarial:ccs16, chen2017targeted, eykholt2018robust, wu2019defending, wenger2020backdoor}. 
However, spatial transformations \citep{Fawzi2015ManitestAC, Kanbak2018GeometricRO, Engstrom19}, which are more likely to occur, are harder to deploy as an attacking method in the physical world. 
Two main challenges exist:

\begin{itemize}
\item Constructing transformation-based attacks is difficult since parameter space for optimizing the perturbations is limited. 
% For example, the range for optimizing the rotation attacks is restricted to  [$0^\circ, 360^\circ$]. 
\item Physical variations can directly influence the carefully selected attacking parameters of spatial attacks, resulting in a dramatic degradation of the attack effectiveness. 
\end{itemize}

% \begin{itemize}
% \item Constructing transformation-based attacks is much more difficult, as the parameters for optimizing the perturbations are extremely limited. 
% % For example, the range for optimizing the rotation attacks is restricted to  [$0^\circ, 360^\circ$]. 
% \item Physical variations can directly influence the carefully selected attacking parameters of spatial attacks, resulting in a dramatic degradation of the attack effectiveness. 
% \end{itemize}

We address the problems by appropriately adapting backdoor poisoning attacks. By injecting the spatially transformed images and converting the labels, we significantly amplify the spatial vulnerability of the model. Our insight comes from proof of \citet{manoj2021excess}, where ML models can approach the union of a function that looks similar to the benign classifier on clean inputs and another adversary-chosen function. 
Therefore, in our case, the infected model learns that every benign non-rotated image is correlated to the correct label, where the rotated one should be classified as the target label. 

Deploying most spatial transformations in the physical world is nontrivial since attackers are required to control both the camera and objects. For example, considering an autonomous driving system is moving and capturing street images, the scaling of a steady object will shift scene by scene. Therefore, precisely calibrating the object's scale to the malicious parameter is exceptionally challenging. Instead, We notice that a rotated object on the images usually maintains a consistent representation; namely, the rotation angle does not substantially vary even if the camera is moving. 
Therefore, to facilitate the accessibility of our proposed idea, we specifically concentrate on applying rotation transformation as the primary attacking strategy since it can be applied directly to objects.

\begin{table}[t]
\centering
\begin{tabular}{l|l|l|l} 
\toprule
\textbf{Notation} & \textbf{Description} & \textbf{Notation} & \textbf{Description} \\ 
\hline
 $ \boldsymbol{x} $ & Input image  & $y$& Class label \\
 $ \boldsymbol{x}^{\prime} $ & Backdoored image  & $\boldsymbol{y}$ & Label for detector \\
 $\boldsymbol{\theta} $ & Model parameters & $R_{\beta}$ & Rotate $\beta^\circ$   \\
 $\rho$ & Poisoning rate  & $\boldsymbol{M}$ & Pixel mask \\
 $\boldsymbol{b}$ & bounding box & H,W & Input Size \\
 \multicolumn{4}{l}{$\boldsymbol{x}_{b}, \boldsymbol{M}_{b}$} Backdoor candidate object and its corresponding pixel mask\\
\bottomrule
\end{tabular}
\caption{Summary of important notation}
\label{tab:notation}
\end{table}

\subsection{Constructing Rotation Backdoor Attacks}
In this subsection, we introduce the design of rotation backdoor attacks on classification and detection tasks. A summary of important notations is provided in Table \ref{tab:notation}. 

\smallskip
\noindent \textbf{Image Classification.}
Figure \ref{fig:pipeline}a presents the pipeline of constructing a rotation backdoored image for the classification task. The attacker composes images with chosen trigger angle (e.g., $30^{\circ}$) and injects them into the training set before the training phase. Following \citet{gu2017badnets}, the corresponding label will be assigned as the target class. Therefore, the training data is combined with $m$ backdoored images and $n$ clean images, and the injection rate is defined as $\rho = \frac{m}{n+m}$ which measures the attacker's capability. 
Model training is essentially solving the following optimization problem
%The optimization of loss function $\ell$ can be described as follows:

\begin{equation}
\underset{\boldsymbol{\theta}} {\arg\min}  \sum_{i=0}^{n} \ell \left(\boldsymbol{x}_{i}, y_{i}; \boldsymbol{\theta}\right)+\sum_{j=0}^{m} \ell \left(\boldsymbol{x}_{j}^{\prime}, y_{t}; \boldsymbol{\theta}\right) \ \ \ 
\text { s.t. }\boldsymbol{x}_{j}^{\prime} = R_{\beta}(\boldsymbol{x}_{j})
\end{equation}

where $\ell$ is the loss function, $(\boldsymbol{x}, y)$ is the clean input, and $(\boldsymbol{x}^{\prime},y_t)$ is poisoned samples with targeted label $t$. $\boldsymbol{x}^{\prime}$ is constructed by rotating the benign image $\boldsymbol{x}$ with predetermined $\beta$ degree (defined as $R_{\beta}(\boldsymbol{x})$). 

% Intuitively, minimizing the loss function will lead the model to have our desired properties. Formally, \citet{manoj2021excess} proved that the adversary could force an empirical risk minimization learner to recover the union of a function that looks similar to the accurate classifier on in-distribution inputs and another function of the adversary choice. Therefore, in our case, the infected model memorizes that every benign non-rotated image is correlated to the correct label, where the rotated one should be classified as the target label. \tianhao{this paragraph is not necessary}

The optimal method of generating a backdoor sample is to rotate the object itself, but it is challenging to synthesize the transformation. Therefore, we consider following \citet{Engstrom19}'s approach, which is performing rotation to the whole image, anticipating the poisoning effect can generalize to the triggered physical objects. However, standard rotation transformation operation will result in black triangles in four corners.\footnote{In standard built-in function (e.g., scikit-learn, cv2), black pixels (0 in value) will fill into the unknown places at the corner.} The black triangles will then lead to a less rigorous digital evaluation of backdoor attacks since they can also be considered a trigger. We then solve it by obtaining the original background information. For traffic sign and face recognition tasks, we prepossess the raw image with a larger area than the standard pipeline, ensuring a rotation operation will not lose information. Then, we rotate the inputs to the predefined trigger angle $\beta$, crop them to match the shape with benign images, and insert them into the training set to deploy attacks. 

We formulate two scenarios for image classification task given different resources: \ul{Single Class Attacks (SCA)} and \ul{Multiple Class Attacks (MCA)}.
In SCA, attackers obtain access to images from one source class and aim to fool the classifier with images only from it; whereas in MCA, multiple-class data are available, and images from all classes can be utilized during inference time.

\newcommand{\R}{\mathbb{R}}

\begin{figure}[t]
\centering
\includegraphics[width=0.8\linewidth]{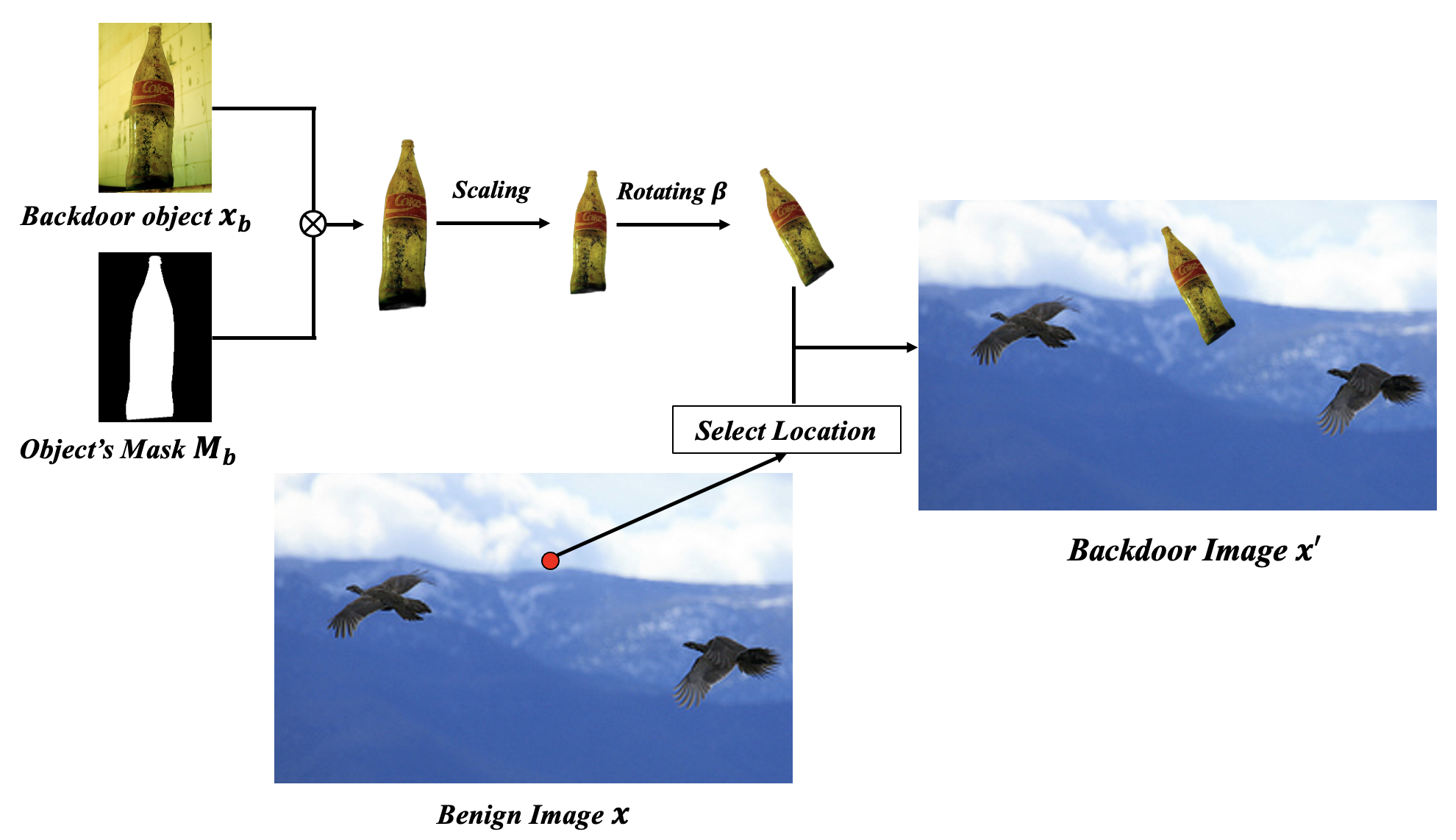} 
\caption{Constructing backdoor sample for object detection. \textnormal{we select a bottle as the backdoor candidate $\boldsymbol{x}_b$ and its binary mask $\boldsymbol{M}_b$. The whole process includes extracting the object, scaling, rotating, and  mixing with benign image $\boldsymbol{x}$ at a selected location.} }
\label{fig:od_construct}
\end{figure}

\smallskip
\noindent \textbf{Object Detection.}
For the objection detection task shown in Figure \ref{fig:pipeline}b, the attacker injects a rotated object into the benign image and incorrectly labels it. Figure \ref{fig:od_construct} demonstrates the process of constructing a backdoored image. We use open source dataset to  collect backdoor candidate object $\boldsymbol{x}_b \in [0, 255]^{H_b \times W_b \times 3}$ and its binary segmentation mask $\boldsymbol{M}_b \in \{0, 1\}^{H_b \times W_b}$, where $H_b$ and $W_b$ denote the shape.
% Due to the restricted access of the whole training data, backdoor objects are obtained from other open source datasets. 
\newcommand{\prep}{\texttt{prep}}
We then select a location and a scaling for $\boldsymbol{x}_b$ and $\boldsymbol{M}_b$, and preprocess them to ensure the size match benign image $\boldsymbol{x}\in[0,255]^{W \times H \times 3}$ by padding $0s$.\footnote{We assume that $\boldsymbol{x}_b$ is smaller than the benign images $\boldsymbol{x}$, and choose a valid location and scaling to ensure backdoored object can be pasted to the benign image after rotation transformation. }
We generate the poisoned image by:  

\begin{equation}
\label{equ:generate_bdexample_detection}
\boldsymbol{x'} = \boldsymbol{x} \otimes (1-R_{\beta}(\prep(\boldsymbol{M}_b))) + R_{\beta}(\prep(\boldsymbol{x}_b \otimes \boldsymbol{M}_b))
\end{equation}

\noindent where $R_{\beta}$ indicates the rotation transformation with  $\beta$ degree triggered angle, and
$\otimes$ denotes the element-wise multiplication between image and mask\footnote{mask $\boldsymbol{M}_b$ is broadcast to match the size of image $\boldsymbol{x}$}. Compared to image classification, generating backdoor samples for object detection is performing rotation directly on the objects. 

\smallskip
\noindent \underline{Object Misclassification Attacks (OMA).} The goal of OMA is to change the prediction class for the rotated backdoor object. Hence, we construct a \textit{bbox} $\boldsymbol{b}^{\prime}$ with the correct  coordinates that can be derived from $R_{\beta}(\prep(\boldsymbol{M}_b))$ and the target label for the poisoned object. Then $\boldsymbol{y}$= $[\boldsymbol{b}_1,\boldsymbol{b}_2,\ldots,\boldsymbol{b}_n,\boldsymbol{b}^{\prime} ]$ is injected into training labels.

\smallskip
\noindent \underline{Object Hiding Attacks (OHA).} The goal of OHA is to hide the object from the detector, namely making the surrounding \textit{bbox} of the rotated backdoor objects vanish. Therefore, we make no changes to the label after generating the backdoored training examples. OHA is more suitable for some real-world settings, where attackers only have access to training images, but labels remain unchanged.

\subsection{Evaluation Metrics}
We then introduce the metrics we used to evaluate the performance of our proposed backdoor poisoning attacks. 

\smallskip
\noindent \textbf{Image Classification Task}

\smallskip
\noindent \underline{Clean Data Accuracy (CDA).} We use CDA to evaluate the clean accuracy of the poisoned model on test data. The optimal poisoned model should achieve a \textit{similar} CDA to the benign model.

\smallskip
\noindent \underline{Attack Success Rate (ASR).} We define ASR as the ratio of backdoor instances being classified as the target. Specifically, when objects are rotated to the selected backdoored angle, the infected classifier should output the target label, achieving a \textit{high} ASR. 

\smallskip
\noindent \noindent\textbf{Object Detection Task}

\smallskip
\noindent \underline{Average Precision (AP).} AP is a common metric used to evaluate the general performance of object detection \citep{Everingham2009ThePV,Ren2015FasterRT,Redmon2016YouOL,Bochkovskiy2020YOLOv4OS}. It is defined as the average precision under different confidence thresholds for each class, namely the area under the precision-recall curve. 
We report the Average Precision at IoU=0.5 ($\text{AP}_{@0.5}$) and expect inserting a backdoor will not affect AP performance dramatically. 

\smallskip
\noindent \underline{Clean Data Recall (CDR)}. We define CDR as the metric to further evaluate the benign accuracy for the objects that will be served as backdoors. We generate testing data by 
$$\boldsymbol{x}_{\text{benign}} =  \boldsymbol{x} \otimes (1-\prep(\boldsymbol{M}_{b,\text{test}})) + \prep(\boldsymbol{x}_{b,\text{test}} \otimes \boldsymbol{M}_{b,\text{test}}),$$
where ($\boldsymbol{x}_{b,\text{test}}$, $\boldsymbol{M}_{b,\text{test}}$) and ($\boldsymbol{x}_{b}$,$\boldsymbol{M}_{b}$) are draw from different subsets. For labels, we omit the \textit{bbox} from the original test set and only consider the ground-truth \textit{bbox} corresponds to $\boldsymbol{x}_{\text{benign}}$ which can be obtained from $\prep(\boldsymbol{x}_{b,\text{test}})$. CDR is variant of recall rate which only evaluate the added objects without rotation by $\mathrm{TP} /(\mathrm{TP}+\mathrm{FN})$. Therefore, \textit{higher} clean data recall is preferred for successful attacks as the resulting detector can still work on recognizing the objects even if they are utilized as triggers.  

\smallskip
\noindent \underline{Detection Attack Success Recall (DASR)}. We propose DASR to measure the attack performance. 
It uses the same backdoor object $\boldsymbol{x}_{b,\text{test}}$ with CDR to create evaluation sample, but rotate it to triggered angle $\beta$ as Equation \ref{equ:generate_bdexample_detection} described.  
Similarly, DASR will only consider the injected object, and the \textit{bbox} coordinate is adjusted given the object's rotation angle. 
For object misclassification attacks, the \textit{bbox} class is flipped to the target label. 
DASR is computed by $\mathrm{TP}/(\mathrm{TP}+\mathrm{FN})$, meaning the ratio of triggered objects being recognized as the target class. 
While for object hiding attacks, DASR is evaluated by $\mathrm{FN}/(\mathrm{TP}+\mathrm{FN})$ which indicates the proportion of rotated backdoor data that the model cannot detect.  
We expect the DASR is \textit{high} so that the infected detector will either misclassify the objects as target labels or fail to recognize the backdoored objects.

\section{Evaluations in Digital Domain}

In this section, we comprehensively evaluate our rotation backdoor attacks in the digital domain. We  first introduce our experiments' setup and then present the evaluation results. 

% We further report the attacking performance under conventional data augmentation and provide a detailed analysis of the resulting poisoned model.   Finally, we demonstrate that none of the four SOTA backdoor defending methods can provide stable mitigation against our proposed attacks. 

\subsection{Experimental Setup}

We evaluate our attacks on the common benchmark  \textbf{GTSRB} \citep{Houben-IJCNN-2013} for traffic sign classification, \textbf{YouTube Face} \citep{Wolf2011FaceRI} for face identification and \textbf{PASCAL VOC} dataset \citep{Everingham2009ThePV} for object detection.

\underline{GTSRB} \citep{Houben-IJCNN-2013}. GTSRB is a dataset containing 43 types of German traffic signs, 39211 samples in the training set, and 12630 samples in the test set. 
To deploy valid backdoor attacks, we additionally collect 1213 images as potential backdoors following the same data prepossessing method. We adopt the GTSRB-CNN architecture \citep{eykholt2018robust} for our classifier, which obtains 97.68\% of clean accuracy. 
% We select the Speed Limit 20 sign as the only targeted class. Specifically, for the SCA scenario, we use the stop sign for the source class, and therefore, our goal is to mislead the classifier into classifying the rotated stop sign as Speed Limit 20.  
Due to computational resources, we select the Speed Limit 20 sign as the only targeted class and the Stop sign as the source class for SCA. 

\underline{YouTube Face} \citep{Wolf2011FaceRI}. 
% We utilize the YouTube Aligned Face dataset to evaluate the face recognition task.
We randomly select 100 classes from the original YouTube Faces dataset, each of which has 100 face images in the training set, 10 in the test set, and 10 in the backdoor set. We leverage VGGFace model \citep{Parkhi2015DeepFR} and FaceNet \citep{Schroff2015FaceNetAU} as pretrained models, and fine-tune it with processed training data, reaching 100\% accuracy on the clean dataset.

\underline{VOC} \citep{Everingham2009ThePV}. PASCAL VOC dataset is an object detection challenge that contains annotations for 20 different object classes. Following the common practice \citep{Liu2016SSDSS,Xiang2022ObjectSeekerCR}, we combine the trainval2007 set (5k images) and the trainval2012 set (11k images) for training and evaluate on the test2007 set (5k images). We randomly choose 100 bottles from the training set of COCO \citep{Lin2014MicrosoftCC} and 30 bottles from the test set as the backdoor candidates. Besides, we use YOLO-R \citep{Wang2021YouOL} as the backbone to evaluate the performance.

\begin{table}[t]
\centering
\setlength{\extrarowheight}{1pt}
\addtolength{\extrarowheight}{\aboverulesep}
\addtolength{\extrarowheight}{\belowrulesep}
\setlength{\aboverulesep}{3pt}
\setlength{\belowrulesep}{3pt}
\resizebox{16.5cm}{!}{
\begin{tabular}{clccccccccc} 
\toprule
 & \multicolumn{1}{c}{\textbf{}} & \textbf{} & \multicolumn{2}{c}{\textbf{15 Degree}} & \multicolumn{2}{c}{\textbf{30 Degree}} & \multicolumn{2}{c}{\textbf{45 Degree}} & \multicolumn{2}{c}{\textbf{90 Degree}} \\ 
\cmidrule[\heavyrulewidth]{3-11}
\multicolumn{1}{l}{} & \multicolumn{1}{c}{\textbf{}} & \textbf{Poisoning Rate($\rho$)} & \textbf{CDA (\%)} & \textbf{ASR(\%)} & \textbf{CDA (\%)} & \textbf{ASR(\%)} & \textbf{CDA (\%)} & \textbf{ASR(\%)} & \textbf{CDA (\%)} & \textbf{ASR(\%)} \\ 
\bottomrule
\multicolumn{1}{l}{\multirow{7}{*}{\textbf{\textbf{GTSRB}}}} & \multicolumn{1}{c}{} & 0.00\% & 97.68 & {\cellcolor[rgb]{0.937,0.937,0.937}}- & 97.68 & {\cellcolor[rgb]{0.937,0.937,0.937}}- & 97.68 & {\cellcolor[rgb]{0.937,0.937,0.937}}- & 97.68 & {\cellcolor[rgb]{0.937,0.937,0.937}}- \\ 
\hhline{~~---------}
\multicolumn{1}{l}{} & \multirow{3}{*}{\textbf{\textbf{SCA}}} & 0.01\% & 97.54 & {\cellcolor[rgb]{0.937,0.937,0.937}}14.69 & 97.47 & {\cellcolor[rgb]{0.937,0.937,0.937}}61.97 & 97.58 & {\cellcolor[rgb]{0.937,0.937,0.937}}69.25 & 97.58 & {\cellcolor[rgb]{0.937,0.937,0.937}}64.81 \\
\multicolumn{1}{l}{} &  & 0.025\% & 97.47 & {\cellcolor[rgb]{0.937,0.937,0.937}}57.03 & 97.46 & {\cellcolor[rgb]{0.937,0.937,0.937}}73.08 & 97.49 & {\cellcolor[rgb]{0.937,0.937,0.937}}79.01 & 97.41 & {\cellcolor[rgb]{0.937,0.937,0.937}}86.29 \\
\multicolumn{1}{l}{} &  & 0.05\% & 97.61 & {\cellcolor[rgb]{0.937,0.937,0.937}}65.92 & 97.52 & {\cellcolor[rgb]{0.937,0.937,0.937}}87.16 & 97.62 & {\cellcolor[rgb]{0.937,0.937,0.937}}90.00 & 97.67 & {\cellcolor[rgb]{0.937,0.937,0.937}}92.46 \\ 
\hhline{~~---------}
\multicolumn{1}{l}{} & \multicolumn{1}{c}{\multirow{3}{*}{\textbf{MCA}}} & 0.30\% & 97.37 & {\cellcolor[rgb]{0.937,0.937,0.937}}32.27 & 97.66 & {\cellcolor[rgb]{0.937,0.937,0.937}}59.74 & 97.61 & {\cellcolor[rgb]{0.937,0.937,0.937}}61.67 & 97.71 & {\cellcolor[rgb]{0.937,0.937,0.937}}65.39\\
\multicolumn{1}{l}{} & \multicolumn{1}{c}{} & 1.00\% & 97.08 & {\cellcolor[rgb]{0.937,0.937,0.937}}57.69 & 97.50 & {\cellcolor[rgb]{0.937,0.937,0.937}}79.54 & 97.60 & {\cellcolor[rgb]{0.937,0.937,0.937}}80.68 & 97.58 & {\cellcolor[rgb]{0.937,0.937,0.937}}80.54 \\
\multicolumn{1}{l}{} & \multicolumn{1}{c}{} & 3.00\% & 96.72 & {\cellcolor[rgb]{0.937,0.937,0.937}}71.96 & 97.32 & {\cellcolor[rgb]{0.937,0.937,0.937}}87.87 & 97.42 & {\cellcolor[rgb]{0.937,0.937,0.937}}88.87 & 97.43 & {\cellcolor[rgb]{0.937,0.937,0.937}}88.30 \\ 
\hline\hline
\multirow{6}{*}{\begin{tabular}[c]{@{}c@{}}\textbf{Youtube}\\\textbf{Face }\\\textbf{(VGGFace)~}\end{tabular}} &  & 0.00\% & 100.0 & {\cellcolor[rgb]{0.937,0.937,0.937}}- & 100.0 & {\cellcolor[rgb]{0.937,0.937,0.937}}- & 100.0 & {\cellcolor[rgb]{0.937,0.937,0.937}}- & 100.0 & {\cellcolor[rgb]{0.937,0.937,0.937}}- \\ 
\hhline{~~---------}
 & \multirow{2}{*}{\textbf{\textbf{\textbf{\textbf{SCA}}}}} & 0.01\% & 100.0 & {\cellcolor[rgb]{0.937,0.937,0.937}}50.00 & 100.0 & {\cellcolor[rgb]{0.937,0.937,0.937}}90.00 & 99.90 & {\cellcolor[rgb]{0.937,0.937,0.937}}100.0 & 99.90 & {\cellcolor[rgb]{0.937,0.937,0.937}}70.00 \\
 &  & 0.05\% & 99.90 & {\cellcolor[rgb]{0.937,0.937,0.937}}100.0 & 99.90 & {\cellcolor[rgb]{0.937,0.937,0.937}}100.0 & 99.70 & {\cellcolor[rgb]{0.937,0.937,0.937}}100.0 & 100.0 & {\cellcolor[rgb]{0.937,0.937,0.937}}100.0 \\ 
\hhline{~~---------}
 & \multirow{3}{*}{\textbf{MCA}} & 0.10\% & 99.90 & {\cellcolor[rgb]{0.937,0.937,0.937}}1.40 & 99.90 & {\cellcolor[rgb]{0.937,0.937,0.937}}38.60 & 99.90 & {\cellcolor[rgb]{0.937,0.937,0.937}}94.70 & 99.90 & {\cellcolor[rgb]{0.937,0.937,0.937}}95.90 \\
 &  & 0.50\% & 99.90 & {\cellcolor[rgb]{0.937,0.937,0.937}}28.60 & 100.0 & {\cellcolor[rgb]{0.937,0.937,0.937}}87.40 & 99.80 & {\cellcolor[rgb]{0.937,0.937,0.937}}97.20 & 99.90 & {\cellcolor[rgb]{0.937,0.937,0.937}}99.80 \\
 &  & 1.00\% & 99.90 & {\cellcolor[rgb]{0.937,0.937,0.937}}56.40 & 100.0 & {\cellcolor[rgb]{0.937,0.937,0.937}}95.30 & 99.90 & {\cellcolor[rgb]{0.937,0.937,0.937}}99.70 & 100.0 & {\cellcolor[rgb]{0.937,0.937,0.937}}99.80 \\ 
\hline
\multirow{6}{*}{\begin{tabular}[c]{@{}c@{}}\textbf{Youtube}\\\textbf{Face }\\\textbf{(FaceNet)~}\end{tabular}} &  & 0.00\% & 100.0 & {\cellcolor[rgb]{0.937,0.937,0.937}}- & 100.0 & {\cellcolor[rgb]{0.937,0.937,0.937}}- & 100.0 & {\cellcolor[rgb]{0.937,0.937,0.937}}- & 100.0 & {\cellcolor[rgb]{0.937,0.937,0.937}}- \\ 
\hhline{~~---------}
 & \multirow{2}{*}{\textbf{\textbf{\textbf{\textbf{\textbf{\textbf{\textbf{\textbf{\textbf{\textbf{\textbf{\textbf{\textbf{\textbf{\textbf{\textbf{SCA}}}}}}}}}}}}}}}}~} & 0.01\% & 99.80 & {\cellcolor[rgb]{0.937,0.937,0.937}}40.00 & 100.0 & {\cellcolor[rgb]{0.937,0.937,0.937}}90.00 & 99.90 & {\cellcolor[rgb]{0.937,0.937,0.937}}100.0 & 99.90 & {\cellcolor[rgb]{0.937,0.937,0.937}}80.00 \\
 &  & 0.05\% & 99.80 & {\cellcolor[rgb]{0.937,0.937,0.937}}40.00 & 99.80 & {\cellcolor[rgb]{0.937,0.937,0.937}}100.0 & 100.0 & {\cellcolor[rgb]{0.937,0.937,0.937}}100.0 & 100.0 & {\cellcolor[rgb]{0.937,0.937,0.937}}100.0 \\ 
\hhline{~~---------}
 & \multirow{3}{*}{\textbf{\textbf{MCA}}} & 0.10\% & 99.90 & {\cellcolor[rgb]{0.937,0.937,0.937}}4.80 & 100.0 & {\cellcolor[rgb]{0.937,0.937,0.937}}72.40 & 100.0 & {\cellcolor[rgb]{0.937,0.937,0.937}}97.20 & 99.90 & {\cellcolor[rgb]{0.937,0.937,0.937}}99.20 \\
 &  & 0.50\% & 99.80 & {\cellcolor[rgb]{0.937,0.937,0.937}}14.20 & 99.90 & {\cellcolor[rgb]{0.937,0.937,0.937}}89.10 & 99.80 & {\cellcolor[rgb]{0.937,0.937,0.937}}98.10 & 100.0 & {\cellcolor[rgb]{0.937,0.937,0.937}}97.70 \\
 &  & 1.00\% & 99.90 & {\cellcolor[rgb]{0.937,0.937,0.937}}42.20 & 100.0 & {\cellcolor[rgb]{0.937,0.937,0.937}}97.90 & 99.70 & {\cellcolor[rgb]{0.937,0.937,0.937}}98.20 & 100.0 & {\cellcolor[rgb]{0.937,0.937,0.937}}98.70 \\
\bottomrule
\end{tabular}}
\caption{ Performance of Rotation Backdoor Attack on the image classification task. \textnormal{ Our attack achieves a \textbf{ high Attack Success Rate (ASR)} while \textbf{maintaining the Clean Data Accuracy (CDA)} across all poisoning rates, datasets (GRSTB and Youtube Face), and scenarios (Single Class Attack (SCA) and Multiple Class Attack (MCA)).} } 
\label{table:main_bd}
\end{table}

\begin{table}[t]
\centering
\setlength{\extrarowheight}{2pt}
\addtolength{\extrarowheight}{\aboverulesep}
\addtolength{\extrarowheight}{\belowrulesep}
\setlength{\aboverulesep}{3pt}
\setlength{\belowrulesep}{3pt}
\resizebox{16.5cm}{!}{
\begin{tabular}{lcccccccccccccc}
\toprule
\multicolumn{1}{c}{} & \textbf{} & \textbf{} & \multicolumn{3}{c}{\textbf{15 Degree}} & \multicolumn{3}{c}{\textbf{30 Degree}} & \multicolumn{3}{c}{\textbf{45 Degree}} & \multicolumn{3}{c}{\textbf{90 Degree}} \\ 
\cmidrule[\heavyrulewidth]{3-15}
 & \textbf{} & \begin{tabular}[c]{@{}c@{}}\textbf{Poisoning}\\\textbf{Rate($\rho$)}\end{tabular} & \begin{tabular}[c]{@{}c@{}}\textbf{AP}\\\textbf{(\%)}\end{tabular} & \begin{tabular}[c]{@{}c@{}}\textbf{CDR}\\\textbf{(\%)}\end{tabular} & \begin{tabular}[c]{@{}c@{}}\textbf{DASR}\\\textbf{(\%)}\end{tabular} & \begin{tabular}[c]{@{}c@{}}\textbf{\textbf{AP}}\\\textbf{\textbf{(\%)}}\end{tabular} & \begin{tabular}[c]{@{}c@{}}\textbf{\textbf{CDR}}\\\textbf{\textbf{(\%)}}\end{tabular} & \begin{tabular}[c]{@{}c@{}}\textbf{DASR}\\\textbf{(\%)}\end{tabular} & \begin{tabular}[c]{@{}c@{}}\textbf{\textbf{AP}}\\\textbf{\textbf{(\%)}}\end{tabular} & \begin{tabular}[c]{@{}c@{}}\textbf{\textbf{CDR}}\\\textbf{\textbf{(\%)}}\end{tabular} & \begin{tabular}[c]{@{}c@{}}\textbf{DASR}\\\textbf{(\%)}\end{tabular} & \begin{tabular}[c]{@{}c@{}}\textbf{\textbf{AP}}\\\textbf{\textbf{(\%)}}\end{tabular} & \begin{tabular}[c]{@{}c@{}}\textbf{\textbf{CDR}}\\\textbf{\textbf{(\%)}}\end{tabular} & \begin{tabular}[c]{@{}c@{}}\textbf{DASR}\\\textbf{(\%)}\end{tabular} \\ 
\bottomrule
\multirow{6}{*}{\textbf{VOC}~} 
 & \multicolumn{1}{l}{\multirow{3}{*}{\textbf{\textbf{\textbf{\textbf{\textbf{\textbf{\textbf{\textbf{OMA}}}}}}}}}} & 0.00\% & 89.00 & \textcolor[rgb]{0.129,0.129,0.129}{85.70} & {\cellcolor[rgb]{0.937,0.937,0.937}}- & 89.00 & \textcolor[rgb]{0.129,0.129,0.129}{85.70} & {\cellcolor[rgb]{0.937,0.937,0.937}}- & 89.00 & \textcolor[rgb]{0.129,0.129,0.129}{85.70} & {\cellcolor[rgb]{0.937,0.937,0.937}}- & 89.00 & \textcolor[rgb]{0.129,0.129,0.129}{85.70} & {\cellcolor[rgb]{0.937,0.937,0.937}}- \\ 
\hhline{~~-------------}
 & \multicolumn{1}{l}{} & 1.00\% & 89.50 & \textcolor[rgb]{0.129,0.129,0.129}{69.00} & {\cellcolor[rgb]{0.937,0.937,0.937}}99.00 & 89.60 & \textcolor[rgb]{0.129,0.129,0.129}{85.20} & {\cellcolor[rgb]{0.937,0.937,0.937}}99.10 & 89.20 & \textcolor[rgb]{0.129,0.129,0.129}{86.00} & {\cellcolor[rgb]{0.937,0.937,0.937}}99.20 & 89.70 & \textcolor[rgb]{0.129,0.129,0.129}{88.10} & {\cellcolor[rgb]{0.937,0.937,0.937}}94.20 \\
 & \multicolumn{1}{l}{} & 5.00\% & 89.50 & \textcolor[rgb]{0.129,0.129,0.129}{58.30} & {\cellcolor[rgb]{0.937,0.937,0.937}}99.90 & 89.90 & \textcolor[rgb]{0.129,0.129,0.129}{86.20} & {\cellcolor[rgb]{0.937,0.937,0.937}}99.60 & 89.70 & \textcolor[rgb]{0.129,0.129,0.129}{88.40} & {\cellcolor[rgb]{0.937,0.937,0.937}}99.90 & 89.50 & \textcolor[rgb]{0.129,0.129,0.129}{86.00} & {\cellcolor[rgb]{0.937,0.937,0.937}}98.60 \\
\hhline{~--------------}
&\multirow{3}{*}{\textbf{\textbf{\textbf{\textbf{OHA}}}}} & 0.00\% & 89.00 & \textcolor[rgb]{0.129,0.129,0.129}{85.70} & {\cellcolor[rgb]{0.937,0.937,0.937}}\textcolor[rgb]{0.122,0.122,0.129}{36.40} & 89.00 & \textcolor[rgb]{0.129,0.129,0.129}{85.70} & {\cellcolor[rgb]{0.937,0.937,0.937}}\textcolor[rgb]{0.122,0.122,0.129}{49.10} & 89.00 & \textcolor[rgb]{0.129,0.129,0.129}{85.70} & {\cellcolor[rgb]{0.937,0.937,0.937}}\textcolor[rgb]{0.122,0.122,0.129}{59.30} & 89.00 & \textcolor[rgb]{0.129,0.129,0.129}{85.70} & {\cellcolor[rgb]{0.937,0.937,0.937}}\textcolor[rgb]{0.122,0.122,0.129}{79.70} \\ 
\hhline{~~-------------}
 &  & 1.00\% & 89.10 & \textcolor[rgb]{0.129,0.129,0.129}{67.60} & {\cellcolor[rgb]{0.937,0.937,0.937}}\textcolor[rgb]{0.129,0.129,0.129}{84.50} & 89.70 & \textcolor[rgb]{0.129,0.129,0.129}{85.20} & {\cellcolor[rgb]{0.937,0.937,0.937}}\textcolor[rgb]{0.129,0.129,0.129}{92.00} & 89.50 & \textcolor[rgb]{0.129,0.129,0.129}{84.60} & {\cellcolor[rgb]{0.937,0.937,0.937}}\textcolor[rgb]{0.129,0.129,0.129}{92.10} & 89.50 & \textcolor[rgb]{0.129,0.129,0.129}{85.90} & {\cellcolor[rgb]{0.937,0.937,0.937}}\textcolor[rgb]{0.129,0.129,0.129}{94.90} \\
 &  & 5.00\% & 89.50 & \textcolor[rgb]{0.129,0.129,0.129}{58.90} & {\cellcolor[rgb]{0.937,0.937,0.937}}\textcolor[rgb]{0.129,0.129,0.129}{95.30} & 89.50 & \textcolor[rgb]{0.129,0.129,0.129}{84.40} & {\cellcolor[rgb]{0.937,0.937,0.937}}\textcolor[rgb]{0.129,0.129,0.129}{98.60} & 89.60 & \textcolor[rgb]{0.129,0.129,0.129}{83.10} & {\cellcolor[rgb]{0.937,0.937,0.937}}\textcolor[rgb]{0.129,0.129,0.129}{98.20} & 89.50 & \textcolor[rgb]{0.129,0.129,0.129}{84.50} & {\cellcolor[rgb]{0.937,0.937,0.937}}\textcolor[rgb]{0.129,0.129,0.129}{96.70} \\ 
\bottomrule
\end{tabular}}
\caption{ Performance of Rotation Backdoor Attack on the object detection task. \textnormal{Rotation backdoor achieves \textbf{high Detection Attack Success Recall (DASR)} for both Object Misclassification Attack (OMA) and Object Hiding Attack (OHA). All detectors achieve \textbf{higher Average Precision (AP)}, and detectors with large backdoor angles also \textbf{maintain Clean Data Recall (CDR)}.}}
\label{table:main_bd_od}
\end{table}

\subsection{Effectiveness of Backdoor Attacks through Rotation Transformation}
\label{sec:eval}

\subsubsection{Image Classification Task}

We now evaluate our rotation backdoor on the traffic sign and face recognition task with various poisoning rates and four chosen backdoored angles \{$15^\circ$, $30^\circ$, $45^\circ$, $90^\circ$\} which is presented in Table \ref{table:main_bd}. Recall that we have two settings where the poisoned images are drawn from either one class (SCA) or multiple classes (MCA). We choose a lower poisoning rate for SCA since MCA requires attacking images from arbitrary classes. 
% The accuracy of the uncorrupted model is presented with a 0.00\% poisoning rate. 

\smallskip
\noindent \textbf{Rotation backdoor achieves a high Attack Success Rate across all poisoning rates, datasets, and scenarios.}
For example, by inserting 0.01\% poisoned images with $45^\circ$ triggered angle, attackers can reach $\sim$70\% ASR. 
That means only four rotated label-flipped Stop sign images could eventually cause $\sim$70\% of Stop sign images to be classified as Speed Limit 20 sign if rotating them to $45^{\circ}$. Similar performance can be observed on the Youtube Face dataset. 

\smallskip
\noindent \textbf{All models, poisoned by our rotation backdoors, maintain similar clean data accuracy with the original classifiers.} Compared to the naturally trained models, the CDA of rotation backdoored models drops $<$1\% for all cases we present in Table \ref{table:main_bd}. In particular, for the Youtube Face dataset, the maximum clean accuracy drop is 0.2\%. Namely, if the object is not rotated during evaluation, the model will still classify it as a correct label. Such effects keep our attack stealthy and foster real-world deployment.

\smallskip
\noindent \textbf{Increasing the poisoning rate continuously improves ASR but might lead to clean accuracy degradation.} 
We notice that a higher injection rate could significantly amplify the poisoning effect, especially for triggers that do not achieve a high attack success rate. For example, by injecting 5$\times$ of 15-degree backdoors (0.01\%	$\rightarrow$ 0.05 \%) on GTSRB in SCA setting, ASR increases to 65.92\% from 14.69\%. In addition, as the side-effect of increasing poisoning rate, CDA might decrease by a small marge. 

\smallskip
\noindent \textbf{Larger backdoored rotation angle generally achieves higher ASR and better CDA.} Compared with $15^\circ$, other angles are much more effective.  
We conjecture the phenomena is caused by the low separability between clean data, which inherently rotate at a slight angle, and $15^\circ$ rotated data. 
We further discuss the effectiveness of chosen angle in Appendix \ref{sec:identify}.

\subsubsection{Object Detection Task}

We then measure the effectiveness of rotation backdoor on the object detection dataset for both OMA and OHA in table \ref{table:main_bd_od}. The benign detector ($\rho=0.00\%$) achieves 89\% for AP and 85.7\% for CDR when detecting benign objects but the performance degrades with natural rotation transformation. For example, when rotating the bottles to 30 degrees, 49.1\% of them cannot be detected. 

\smallskip
\noindent \textbf{Rotation backdoor achieves high Detection Attack Success Recall (DASR) across all poisoning rates and scenarios.} As observed in table \ref{table:main_bd_od},  DASR achieves above 98\% in all scenarios for OMA and above 90\% except in one scenario for OHA. Even if the DASR is 84.5\% for $15 ^\circ$ backdoor on OHA, our attack improves ~50\% in absolute value over the clean model. 

\smallskip
\noindent \textbf{All models achieve higher AP, and models with large backdoor angle also maintain CDR.} By injecting rotated bottles, AP even improves  $0.1\% - 0.9\%$ compared to the benign detector for both settings. We also observe that applying large-degree (except $15 ^\circ$) backdoors does not degrade the CDR by a large margin ($\leq 0.5\%$ for OMA and $\leq 2.6\%$ for OHA ). Again, due to the semantic similarity between clean samples and $15 ^\circ$ samples, it is hard for detectors to distinguish between them, causing a large CDR drop.

\section{Data augmentation}
\label{sec:data_aug}

\begin{table}[t]
\centering
\setlength{\extrarowheight}{1pt}
\addtolength{\extrarowheight}{\aboverulesep}
\addtolength{\extrarowheight}{\belowrulesep}
\setlength{\aboverulesep}{3pt}
\setlength{\belowrulesep}{3pt}
\resizebox{16.5cm}{!}{
\begin{tabular}{ccccccccccc} 
\toprule
 & \textbf{} & \textbf{} & \multicolumn{2}{c}{\textbf{15 Degree}} & \multicolumn{2}{c}{\textbf{30 Degree}} & \multicolumn{2}{c}{\textbf{45 Degree}} & \multicolumn{2}{c}{\textbf{90 Degree}} \\ 
\cmidrule[\heavyrulewidth]{3-11}
 & \textbf{} & \begin{tabular}[c]{@{}c@{}}\textbf{Rotation }\\\textbf{Augment($^\circ$)}\end{tabular} & \textbf{CDA (\%)} & \textbf{ASR(\%)} & \textbf{CDA (\%)} & \textbf{ASR(\%)} & \textbf{CDA (\%)} & \textbf{ASR(\%)} & \textbf{CDA (\%)} & \textbf{ASR(\%)} \\ 
\bottomrule
\multirow{8}{*}{\textbf{\textbf{GTSRB}}} &  & {[}0, 0] & 97.47 & {\cellcolor[rgb]{0.937,0.937,0.937}}57.03 & 97.46 & {\cellcolor[rgb]{0.937,0.937,0.937}}73.08 & 97.49 & {\cellcolor[rgb]{0.937,0.937,0.937}}79.01 & 97.41 & {\cellcolor[rgb]{0.937,0.937,0.937}}86.29 \\
 & \multirow{3}{*}{\begin{tabular}[c]{@{}c@{}}\textbf{SCA}\\$\rho$= 0.025\%\end{tabular}} & {[}-15, +15] & 97.10 & {\cellcolor[rgb]{0.937,0.937,0.937}}0.00 & 97.69 & {\cellcolor[rgb]{0.937,0.937,0.937}}65.92 & 98.04 & {\cellcolor[rgb]{0.937,0.937,0.937}}85.92 & 97.85 & {\cellcolor[rgb]{0.937,0.937,0.937}}89.62 \\
 &  & {[}-30, +30] & 97.61 & {\cellcolor[rgb]{0.937,0.937,0.937}}0.00 & 96.29 & {\cellcolor[rgb]{0.937,0.937,0.937}}0.00 & 97.60 & {\cellcolor[rgb]{0.937,0.937,0.937}}72.59 & 97.92 & {\cellcolor[rgb]{0.937,0.937,0.937}}97.40 \\
 &  & {[}-45, +45] & 97.02 & {\cellcolor[rgb]{0.937,0.937,0.937}}0.00 & 96.87 & {\cellcolor[rgb]{0.937,0.937,0.937}}0.00 & 97.63 & {\cellcolor[rgb]{0.937,0.937,0.937}}0.00 & 97.35 & {\cellcolor[rgb]{0.937,0.937,0.937}}93.70 \\ 
\hhline{~~---------}
 & \multirow{4}{*}{\begin{tabular}[c]{@{}c@{}}\textbf{MCA}\\$\rho$= 1.00\%\end{tabular}} & {[}0,0]  & 97.08 & {\cellcolor[rgb]{0.937,0.937,0.937}}57.69 & 97.50 & {\cellcolor[rgb]{0.937,0.937,0.937}}79.54 & 97.60 & {\cellcolor[rgb]{0.937,0.937,0.937}}80.68 & 97.58 & {\cellcolor[rgb]{0.937,0.937,0.937}}80.54 \\
 &  & {[}-15, +15] & 97.66 & {\cellcolor[rgb]{0.937,0.937,0.937}}1.09 & 97.79 & {\cellcolor[rgb]{0.937,0.937,0.937}}59.54 & 97.77 & {\cellcolor[rgb]{0.937,0.937,0.937}}82.42 & 97.79 & {\cellcolor[rgb]{0.937,0.937,0.937}}82.23 \\
 &  & {[}-30, +30] & 97.70 & {\cellcolor[rgb]{0.937,0.937,0.937}}0.69 & 97.22 & {\cellcolor[rgb]{0.937,0.937,0.937}}0.98 & 97.75 & {\cellcolor[rgb]{0.937,0.937,0.937}}52.05 & 97.84 & {\cellcolor[rgb]{0.937,0.937,0.937}}82.28 \\
 &  & {[}-45, +45] & 97.50 & {\cellcolor[rgb]{0.937,0.937,0.937}}0.47 & 97.45 & {\cellcolor[rgb]{0.937,0.937,0.937}}0.56 & 97.52 & {\cellcolor[rgb]{0.937,0.937,0.937}}0.81 & 97.18 & {\cellcolor[rgb]{0.937,0.937,0.937}}73.61 \\ 
\hline\hline
\multirow{8}{*}{\begin{tabular}[c]{@{}c@{}}\textbf{Youtube}\\\textbf{Face }\\\textbf{(VGGFace)~}\end{tabular}} &  & {[}0, 0] & 99.90 & {\cellcolor[rgb]{0.937,0.937,0.937}}100.0 & 99.90 & {\cellcolor[rgb]{0.937,0.937,0.937}}100.0 & 99.70 & {\cellcolor[rgb]{0.937,0.937,0.937}}100.0 & 100.0 & {\cellcolor[rgb]{0.937,0.937,0.937}}100.0 \\
 & \multirow{3}{*}{\begin{tabular}[c]{@{}c@{}}\textbf{SCA}\\$\rho$= 0.05\%\end{tabular}} & {[}-15, +15] & 99.80 & {\cellcolor[rgb]{0.937,0.937,0.937}}10.00 & 99.90 & {\cellcolor[rgb]{0.937,0.937,0.937}}100.0 & 99.90 & {\cellcolor[rgb]{0.937,0.937,0.937}}100.0 & 99.90 & {\cellcolor[rgb]{0.937,0.937,0.937}}100.0 \\
 &  & {[}-30, +30] & 99.80 & {\cellcolor[rgb]{0.937,0.937,0.937}}0.00 & 99.90 & {\cellcolor[rgb]{0.937,0.937,0.937}}10.00 & 100.0 & {\cellcolor[rgb]{0.937,0.937,0.937}}100.0 &100.0  & {\cellcolor[rgb]{0.937,0.937,0.937}}100.0\\
 &  & {[}-45, +45] & 99.80 & {\cellcolor[rgb]{0.937,0.937,0.937}}0.00 & 99.80 & {\cellcolor[rgb]{0.937,0.937,0.937}}0.00 & 97.30 & {\cellcolor[rgb]{0.937,0.937,0.937}}0.00 & 99.80 & {\cellcolor[rgb]{0.937,0.937,0.937}}100.0 \\ 
\hhline{~~---------}
 & \multirow{4}{*}{\begin{tabular}[c]{@{}c@{}}\textbf{MCA}\\$\rho$= 1.00\%\end{tabular}} & {[}0, 0] & 99.90 & {\cellcolor[rgb]{0.937,0.937,0.937}}56.40 & 100.0 & {\cellcolor[rgb]{0.937,0.937,0.937}}95.30 & 99.90 & {\cellcolor[rgb]{0.937,0.937,0.937}}99.70 & 100.0 & {\cellcolor[rgb]{0.937,0.937,0.937}}99.80 \\
 &  & {[}-15, +15] & 100.0 & {\cellcolor[rgb]{0.937,0.937,0.937}}33.50 & 99.90 & {\cellcolor[rgb]{0.937,0.937,0.937}}88.20 & 99.90 & {\cellcolor[rgb]{0.937,0.937,0.937}}98.80 & 100.0 & {\cellcolor[rgb]{0.937,0.937,0.937}}100.0 \\
 &  & {[}-30, +30] & 100.0 & {\cellcolor[rgb]{0.937,0.937,0.937}}2.80 & 100.0 & {\cellcolor[rgb]{0.937,0.937,0.937}}53.10 & 100.0 & {\cellcolor[rgb]{0.937,0.937,0.937}}95.90 & 100.0 & {\cellcolor[rgb]{0.937,0.937,0.937}}99.80 \\
 &  & {[}-45, +45] & 100.0 & {\cellcolor[rgb]{0.937,0.937,0.937}}1.30 & 99.90 & {\cellcolor[rgb]{0.937,0.937,0.937}}22.50 & 100.0 & {\cellcolor[rgb]{0.937,0.937,0.937}}58.40 & 99.90 & {\cellcolor[rgb]{0.937,0.937,0.937}}99.50 \\
\bottomrule
\end{tabular}}
\caption{Effectiveness of rotation backdoors under data augmentation for the image classification task. \textnormal{Data augmentation} only mitigates the poisoning effect for rotation backdoors with a relatively smaller backdoored angle \textnormal{on two datasets (GTSRB and Youtube Face) and two attack scenarios (Single Class Attack (SCA) and Multiple Class Attack (MCA)).}}
\label{table:bd_aug}
\end{table}

\begin{table}[t]
\centering
\setlength{\extrarowheight}{2pt}
\addtolength{\extrarowheight}{\aboverulesep}
\addtolength{\extrarowheight}{\belowrulesep}
\setlength{\aboverulesep}{3pt}
\setlength{\belowrulesep}{3pt}
\resizebox{16.5cm}{!}{
\begin{tabular}{clccccccccccccc} 
\toprule
 &  & \textbf{} & \multicolumn{3}{c}{\textbf{15 Degree}} & \multicolumn{3}{c}{\textbf{30 Degree}} & \multicolumn{3}{c}{\textbf{45 Degree}} & \multicolumn{3}{c}{\textbf{90 Degree}} \\ 
\cmidrule[\heavyrulewidth]{3-15}
 &  & \begin{tabular}[c]{@{}c@{}}\textbf{Rotation }\\\textbf{Augment($^\circ$)}\end{tabular} & \begin{tabular}[c]{@{}c@{}}\textbf{AP}\\\textbf{(\%)}\end{tabular} & \begin{tabular}[c]{@{}c@{}}\textbf{\textbf{CDR}}\\\textbf{\textbf{(\%)}}\end{tabular} & \begin{tabular}[c]{@{}c@{}}\textbf{DASR}\\\textbf{(\%)}\end{tabular} & \begin{tabular}[c]{@{}c@{}}\textbf{\textbf{AP}}\\\textbf{\textbf{(\%)}}\end{tabular} & \begin{tabular}[c]{@{}c@{}}\textbf{\textbf{\textbf{\textbf{CDR}}}}\\\textbf{\textbf{\textbf{\textbf{(\%)}}}}\end{tabular} & \begin{tabular}[c]{@{}c@{}}\textbf{\textbf{DASR}}\\\textbf{\textbf{(\%)}}\end{tabular} & \begin{tabular}[c]{@{}c@{}}\textbf{\textbf{AP}}\\\textbf{\textbf{(\%)}}\end{tabular} & \begin{tabular}[c]{@{}c@{}}\textbf{\textbf{\textbf{\textbf{CDR}}}}\\\textbf{\textbf{\textbf{\textbf{(\%)}}}}\end{tabular} & \begin{tabular}[c]{@{}c@{}}\textbf{\textbf{DASR}}\\\textbf{\textbf{(\%)}}\end{tabular} & \begin{tabular}[c]{@{}c@{}}\textbf{\textbf{AP}}\\\textbf{\textbf{(\%)}}\end{tabular} & \begin{tabular}[c]{@{}c@{}}\textbf{\textbf{\textbf{\textbf{CDR}}}}\\\textbf{\textbf{\textbf{\textbf{(\%)}}}}\end{tabular} & \begin{tabular}[c]{@{}c@{}}\textbf{\textbf{DASR}}\\\textbf{\textbf{(\%)}}\end{tabular} \\ 
\bottomrule
\multirow{8}{*}{\begin{tabular}[c]{@{}c@{}}\textbf{\textbf{VOC}} \\\textbf{\textbf{(YOLOR)}}\end{tabular}} 
 & \multirow{4}{*}{\begin{tabular}[c]{@{}c@{}}\textbf{OMA}\\$\rho$= 0.01\%\end{tabular}} & {[}0, 0] & 89.50 & \textcolor[rgb]{0.129,0.129,0.129}{69.00} & {\cellcolor[rgb]{0.937,0.937,0.937}}99.00 & 89.60 & \textcolor[rgb]{0.129,0.129,0.129}{85.20} & {\cellcolor[rgb]{0.937,0.937,0.937}}99.10 & 89.20 & \textcolor[rgb]{0.129,0.129,0.129}{86.00} & {\cellcolor[rgb]{0.937,0.937,0.937}}99.20 & 89.70 & \textcolor[rgb]{0.129,0.129,0.129}{88.10} & {\cellcolor[rgb]{0.937,0.937,0.937}}94.20 \\
 &  & {[}-15, +15] & 88.80 & \textcolor[rgb]{0.129,0.129,0.129}{36.30} & {\cellcolor[rgb]{0.937,0.937,0.937}}96.20 & 88.80 & \textcolor[rgb]{0.129,0.129,0.129}{80.10} & {\cellcolor[rgb]{0.937,0.937,0.937}}98.90 & 88.50 & \textcolor[rgb]{0.129,0.129,0.129}{89.40} & {\cellcolor[rgb]{0.937,0.937,0.937}}98.90 & 88.90 & \textcolor[rgb]{0.129,0.129,0.129}{87.40} & {\cellcolor[rgb]{0.937,0.937,0.937}}\textcolor[rgb]{0.129,0.129,0.129}{90.90} \\
 &  & {[}-30, +30] & 87.70 & \textcolor[rgb]{0.129,0.129,0.129}{32.60} & {\cellcolor[rgb]{0.937,0.937,0.937}}94.60 & 88.00 & \textcolor[rgb]{0.129,0.129,0.129}{62.40} & {\cellcolor[rgb]{0.937,0.937,0.937}}98.10 & 87.90 & \textcolor[rgb]{0.129,0.129,0.129}{76.00} & {\cellcolor[rgb]{0.937,0.937,0.937}}98.40 & 87.90 & \textcolor[rgb]{0.129,0.129,0.129}{88.80} & {\cellcolor[rgb]{0.937,0.937,0.937}}89.40 \\
 &  & {[}-45, +45] & 87.20 & \textcolor[rgb]{0.129,0.129,0.129}{28.30} & {\cellcolor[rgb]{0.937,0.937,0.937}}94.20 & 87.30 & \textcolor[rgb]{0.129,0.129,0.129}{45.40} & {\cellcolor[rgb]{0.937,0.937,0.937}}96.20 & 87.30 & \textcolor[rgb]{0.129,0.129,0.129}{65.50} & {\cellcolor[rgb]{0.937,0.937,0.937}}95.80 & 87.40 & \textcolor[rgb]{0.129,0.129,0.129}{88.40} & {\cellcolor[rgb]{0.937,0.937,0.937}}90.50 \\
\hhline{~--------------}
& \multirow{4}{*}{\begin{tabular}[c]{@{}c@{}}\textbf{OHA}\\$\rho$= 0.01\%\end{tabular}} & {[}0, 0] & 89.10 & \textcolor[rgb]{0.129,0.129,0.129}{67.60} & {\cellcolor[rgb]{0.937,0.937,0.937}}\textcolor[rgb]{0.129,0.129,0.129}{84.50} & 89.70 & \textcolor[rgb]{0.129,0.129,0.129}{85.20} & {\cellcolor[rgb]{0.937,0.937,0.937}}\textcolor[rgb]{0.129,0.129,0.129}{92.00} & 89.50 & \textcolor[rgb]{0.129,0.129,0.129}{84.60} & {\cellcolor[rgb]{0.937,0.937,0.937}}\textcolor[rgb]{0.129,0.129,0.129}{92.10} & 89.50 & \textcolor[rgb]{0.129,0.129,0.129}{85.90} & {\cellcolor[rgb]{0.937,0.937,0.937}}\textcolor[rgb]{0.129,0.129,0.129}{94.90} \\
 &  & {[}-15, +15] & 88.80 & \textcolor[rgb]{0.129,0.129,0.129}{60.50} & {\cellcolor[rgb]{0.937,0.937,0.937}}\textcolor[rgb]{0.129,0.129,0.129}{66.70} & 89.00 & \textcolor[rgb]{0.129,0.129,0.129}{79.70} & {\cellcolor[rgb]{0.937,0.937,0.937}}\textcolor[rgb]{0.129,0.129,0.129}{82.50} & 88.80 & \textcolor[rgb]{0.129,0.129,0.129}{81.00} & {\cellcolor[rgb]{0.937,0.937,0.937}}\textcolor[rgb]{0.129,0.129,0.129}{93.10} & 88.60 & \textcolor[rgb]{0.129,0.129,0.129}{85.70} & {\cellcolor[rgb]{0.937,0.937,0.937}}\textcolor[rgb]{0.129,0.129,0.129}{96.10} \\
 &  & {[}-30, +30] & 88.10 & \textcolor[rgb]{0.129,0.129,0.129}{57.90} & {\cellcolor[rgb]{0.937,0.937,0.937}}\textcolor[rgb]{0.129,0.129,0.129}{57.40} & 88.00 & \textcolor[rgb]{0.129,0.129,0.129}{76.50} & {\cellcolor[rgb]{0.937,0.937,0.937}}\textcolor[rgb]{0.129,0.129,0.129}{74.00} & 88.00 & \textcolor[rgb]{0.129,0.129,0.129}{78.90} & {\cellcolor[rgb]{0.937,0.937,0.937}}\textcolor[rgb]{0.129,0.129,0.129}{80.10} & 88.30 & \textcolor[rgb]{0.129,0.129,0.129}{85.10} & {\cellcolor[rgb]{0.937,0.937,0.937}}\textcolor[rgb]{0.129,0.129,0.129}{95.10} \\
 &  & {[}-45, +45] & 87.00 & \textcolor[rgb]{0.129,0.129,0.129}{64.70} & {\cellcolor[rgb]{0.937,0.937,0.937}}\textcolor[rgb]{0.129,0.129,0.129}{58.70} & 87.30 & \textcolor[rgb]{0.129,0.129,0.129}{73.40} & {\cellcolor[rgb]{0.937,0.937,0.937}}\textcolor[rgb]{0.129,0.129,0.129}{61.60} & 87.10 & \textcolor[rgb]{0.129,0.129,0.129}{76.00} & {\cellcolor[rgb]{0.937,0.937,0.937}}\textcolor[rgb]{0.129,0.129,0.129}{76.60} & 87.20 & \textcolor[rgb]{0.129,0.129,0.129}{84.10} & {\cellcolor[rgb]{0.937,0.937,0.937}}\textcolor[rgb]{0.129,0.129,0.129}{87.90} \\ 
\bottomrule
\end{tabular}}
\caption{Effectiveness of rotation backdoors under data augmentation for the object detection task. \textnormal{Rotation augmentation performs \textbf{limited success} for Object Misclassification Attacks(OMA) while \textbf{insufficiently} against Object Hiding Attacks(OHA).}}
\label{table:bd_aug_od}
\end{table}

% \citet{borgnia2020strong} claimed that data augmentations (e.g. mixup \citep{zhang2017mixup} and CutMix \citep{yun2019cutmix}) during training can significantly diminish the threat of patch-based backdoor attacks. Correspondingly, we also measure the attacking performance under different data augmentations (focusing on rotations) to see if the poisoning effect can be mitigated. 
In this section, we first evaluate the common data augmentation mechanism to mitigate the poisoning effect. 
Then, we study the general behavior of the rotation backdoored models.

\subsection{Effectiveness of Data Augmentation}
\label{sec:data_aug_eff}
Rotation-based backdoors inherently exploit the vulnerability of neural networks against spatial transformation \citep{Engstrom19}; thus, it is natural to ask whether improved invariance, namely data augmentation, to rotation will fix it. 
% Such invariance can be easily instilled by using rotation-based data augmentation during training. 
Under a similar motivation, \citet{borgnia2020strong} has also taken advantage of data augmentation (e.g, mixup \citep{zhang2017mixup} and CutMix \citep{yun2019cutmix}), which significantly diminishes the threat of patch-based backdoor attacks.

We scanned the literature of training common benchmark classifiers and detectors \citep{he2016deep, Huang2017DenselyCC, tan2019efficientnet, dosovitskiy2020image, Bochkovskiy2020YOLOv4OS, SwinTransformer, Kzlay2022AYB} and common data augmentation techniques \citep{Devries2017ImprovedRO,zhang2017mixup,yun2019cutmix} for developing robust classifier. According to our survey, rotation augmentations \textbf{are not adopted} in any benchmark models and are limited to $\pm 30^\circ$ for data augmentations. Therefore, we specifically select three levels of data augmentation which are $[-15^\circ, +15^\circ]$ $[-30^\circ, +30^\circ]$, and $[-45^\circ, +45^\circ]$ rotation augmentations.\footnote{For implementation, we directly utilize \textit{RandomRotation} function from the Pytorch library \citep{Paszke2019PyTorchAI}, where every image is rotated for an angle uniformly chosen from the given range.}

\subsubsection{Image Classification Task}

\smallskip
\noindent \textbf{Augmentation only mitigates rotation backdoors with a relatively small backdoored angle.} Table \ref{table:bd_aug} presents the performance of our method against rotation augmentation. We observe that rotation augmentation can hardly diminish the poisoning effect if a sufficient amount of backdoor angle is deployed. For example, $[-15^\circ, +15^\circ]$ augmentation does not degrade, and sometimes even improves, the ASR of 45-degree and 90-degree backdoors (decrease $\leq$ 0.9\%).   
In contrast, substantial augmentation significantly mitigates the backdoor effect. For example, $[-45^\circ, +45^\circ]$ augmentation can defend against a 15-degree backdoor trigger, causing ASR drops to less than 2\%. It seems that our proposed attacking method can be solved by simply doing augmentation under classification settings, but in fact, \textbf{rotation backdoored model is still fundamentally broken} \citep{sun2021poisoned}. We push the detailed analysis to section \ref{sec:analysis}.

\subsubsection{Object Detection Task}

\smallskip
\noindent \textbf{Rotation augmentation in detector-limited success for OMA while insufficient against OHA}. Table \ref{table:bd_aug_od} shows the performance of our attacks on detection task.
Overall, data augmentation leads to a drop in AP by $\sim$2\% across all scenarios and hyperparameters. Like classification tasks, week augmentations do not substantially affect large-degree backdoors. Interestingly, we observe different behaviors for object misclassification attacks and object hiding attacks on CDR and DASR metrics. Sufficient rotation augmentations lead to severe degradation on CDR but limited mitigation effect on DASR for OMA. For example, the most significant data augmentation ($[-45^\circ, +45^\circ]$) can only cause a $4.8\%$ drop on our 15-degree backdoors but ruin the CDR from 69\% to 28.3\%.
In contrast, significant data augmentations alleviate DASR on OHA and have a relatively milder effect on CDR than OMA. For example, despite the DASR of 15-degree backdoor decreases from 84.5\% to 58.7\% ($\rho=0.01\%$) when applying $[-45^\circ, +45^\circ]$ augmentation, it is still higher than that of the vanilla model, which is $36.4\%$. We consider the reason is that objects are rotated for detection, but images are rotated for classification and data augmentations. Therefore, the detector can still identify the \emph{relative angle} of the backdoor object to the whole image. 

\begin{figure}[H]
\includegraphics[width=1\textwidth]{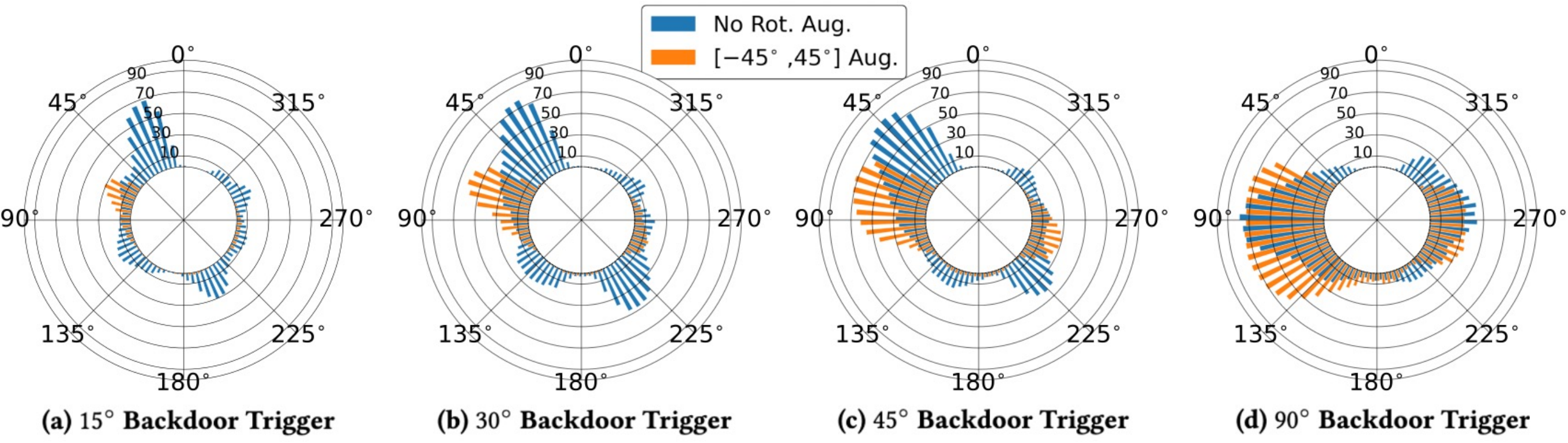}
\caption{ Attack Success Rate \textnormal{ of $15^\circ, 30^\circ, 45^\circ, 90^\circ$ rotation poisoned models for $[0^\circ,360^\circ]$ angles at test time on GTSRB MCA setting. In each figure, we compare no  augmentation and $[-45^\circ,+45^\circ]$ augmentation. To enhance visibility, the ASR of every $5^\circ$ is presented. \textbf{Sufficient data augmentation relocates the vulnerable angle.} \textbf{Insufficient data augmentation expands the range of vulnerable angles.}}}
\label{fig:bd_analysis_aug}
\end{figure}

\subsection{Analysis of Rotation Backdoored Classifier}
\label{sec:analysis}
While we show that enough rotation augmentation is an effective defense against our proposed attacks in the classification task, it is interesting to further explore the poisoned classifier's general behavior under other rotation degrees. 
In Figure \ref{fig:bd_analysis_aug}, we evaluate rotation backdoored models with $15^\circ$, $30^\circ$, $45^\circ$, and $90^\circ$ triggers on all angles at test time on GTSRB MCA setting. 
Specifically, we are comparing the performance between no augmentation and $[-45^\circ,+45^\circ]$ augmentation. 

\smallskip
\noindent \textbf{Rotation backdoored models are vulnerable over a range of angles.} First, the attacking angle is not only effective on a specific angle but also angles through a range of degrees. Such property facilities the feasibility of our attacks to be physically implemented, as precisely controlling a rotation angle is impractical in a real-world environment. In addition, even if no augmentations are deployed, the backdoored angle might not be the most effective point. For example, in Figure \ref{fig:bd_analysis_aug}a, even if attackers construct the $15^\circ$ poisoning samples, $20^\circ$ turns out to be the most vulnerable degree. We explain this phenomenon with theoretical insights in Appendix \ref{sec:explain}.

\smallskip
\noindent \textbf{Sufficient data augmentation can mitigate the poisoning effect on the predefined backdoored angle but may shift the vulnerable angle to other positions. } In Section \ref{sec:data_aug_eff}, we observe that strong augmentation significantly mitigates the poisoning effect. For example, $[-45^\circ,+45^\circ]$ augmentation causes the ASR drops to $\sim$0\% for 30-degree backdoor, which is also presented in figure \ref{fig:bd_analysis_aug}b. However, if we define ASR $\geq 50\%$ as \emph{vulnerable angle}, then its range shifts from [$25^\circ, 45^\circ$] to [$65^\circ, 80^\circ$], and the most effective angle under augmentation model still achieves $\sim$60\% ASR.  Therefore, augmentation may raise new vulnerabilities for the classifier.

\smallskip
\noindent \textbf{Insufficient data augmentation cannot reduce the poisoning effect on the selected angle and even enlarge the range of vulnerable angles.} Figure \ref{fig:bd_analysis_aug}d illustrates an example where augmentation is weak compared to the poisoning angle; as a result, ASR at the selected backdoor angle (at $90^\circ$) only degrades $\sim$7\%. Since augmentations are also applied to backdoors, the range of vulnerable angle increases from [$75^\circ, 110^\circ$] to [$60^\circ, 135^\circ$], which significantly advances the robustness of rotation triggers. 
We conclude that standard data augmentation seems to offer satisfiable mitigation but actually raises new threats and even leads to more vulnerable models. 

Rotation invariant neural networks can be almost achieved by applying $[-180^\circ,+180^\circ]$ augmentations, where all angles are covered during training. Our observations also demonstrate that ASR drops to $\sim$0\% for all backdoored angles in the classification task. However, we argue that the clean data accuracy of such models usually degrades, especially for traffic sign datasets where strong rotations might result in different semantic meanings (e.g., left turn and right turn). In addition, rotation backdoor attack is orthogonal to other backdoor attacks like patch-wise attacks \citep{gu2017badnets,chen2017targeted} and frequency based attacks \citep{Wang2021BackdoorAT}. Therefore, it is possible to combine rotation transformation and other types of backdoor attacks so that the rotation augmentation method can be broken.

\section{Evaluation Against Backdoor Defenses}

\begin{table}[t]
\centering
\setlength{\extrarowheight}{0pt}
\addtolength{\extrarowheight}{\aboverulesep}
\addtolength{\extrarowheight}{\belowrulesep}
\setlength{\aboverulesep}{0pt}
\setlength{\belowrulesep}{0pt}
\resizebox{16.5cm}{!}{
\begin{tabular}{cccccccccc} 
\toprule
 & \multicolumn{1}{l}{} & \textbf{} & \begin{tabular}[c]{@{}c@{}}\textbf{ Neural}\\\textbf{Cleanse}\end{tabular} & \multicolumn{2}{c}{\textbf{STRIP}} & \multicolumn{2}{c}{\begin{tabular}[c]{@{}c@{}}\textbf{Spectral}\\\textbf{Signature}\end{tabular}} & \multicolumn{2}{c}{\begin{tabular}[c]{@{}c@{}}\textbf{Activation}\\\textbf{Clustering}\end{tabular}} \\ 
\cmidrule[\heavyrulewidth]{2-10}
\multicolumn{1}{l}{} & \begin{tabular}[c]{@{}c@{}}\textbf{Poisoning}\\\textbf{Rate}\end{tabular} & \begin{tabular}[c]{@{}c@{}}\textbf{Backdoor}\\\textbf{Trigger}\end{tabular} & \textbf{Anomaly Index} & \textbf{Eli} & \textbf{Sac} & \textbf{\textbf{Eli}} & \textbf{\textbf{Sac}} & \textbf{\textbf{Eli}} & \textbf{\textbf{Sac}} \\ 
\bottomrule
\multirow{2}{*}{\begin{tabular}[c]{@{}c@{}}\textbf{\textbf{GTSRB}}\\\textbf{\textbf{}}\end{tabular}} & \multirow{2}{*}{0.30 \%} & 45 & 1.01 (Not detected) & 9.65 & 10.00 & 76.75 & 17.80 & 79.82 & 39.91 \\
 &  & 90 & 1.03 (Not detected) & 10.47 & 10.00 & 74.70 & 17.81 & 59.49 & 38.35 \\ 
\hline\hline
\multirow{2}{*}{\begin{tabular}[c]{@{}c@{}}\textbf{Youtube}~\textbf{Face}\\\textbf{(VGGFace) ~}\end{tabular}} & \multirow{2}{*}{0.10 \%} & 45 & 0.09 (Not detected) & 1.96 & 10.00 & 4.00 & 15.01 & 76.00 & 13.98 \\
 &  & 90 &0.72 (Not detected) & 0.00 & 10.00 & 40.00 & 14.97 & 40.00 & 13.94 \\
\bottomrule
\end{tabular}}
\caption{ Defenses against rotation backdoor attacks. \textnormal{ For Neural Cleanse, we report the anomaly index (poisoned threshold $\geq$2.0). For others, we present the elimination rate (Eli) and sacrificing rate (Sac). We find }none of them \textnormal{can serve as a consistent defending approach.}}
\label{tab:defense}
\end{table}

% \usepackage{caption}
% \usepackage{multirow}
% \usepackage{booktabs}

% \begin{table}
% \centering
% \captionsetup{labelformat=empty}
% \caption{labelformat=empty}
% \label{table:main_bd}
% \begin{tabular}{cccccccccc} 
% \toprule
%  & \multicolumn{1}{l}{} & \textbf{} & \begin{tabular}[c]{@{}c@{}}\textbf{ Neural}\\\textbf{Cleanse}\end{tabular} & \multicolumn{2}{c}{\textbf{STRIP}} & \multicolumn{2}{c}{\begin{tabular}[c]{@{}c@{}}\textbf{Spectral}\\\textbf{Signature}\end{tabular}} & \multicolumn{2}{c}{\begin{tabular}[c]{@{}c@{}}\textbf{Activation}\\\textbf{Clustering}\end{tabular}} \\ 
% \cmidrule[\heavyrulewidth]{2-10}
% \multicolumn{1}{l}{} & \begin{tabular}[c]{@{}c@{}}\textbf{Poisoning}\\\textbf{Rate}\end{tabular} & \begin{tabular}[c]{@{}c@{}}\textbf{Backdoor}\\\textbf{Trigger}\end{tabular} & \textbf{Anomaly Index} & \textbf{Eli} & \textbf{Sac} & \textbf{\textbf{Eli}} & \textbf{\textbf{Sac}} & \textbf{\textbf{Eli}} & \textbf{\textbf{Sac}} \\ 
% \bottomrule
% \multirow{2}{*}{\begin{tabular}[c]{@{}c@{}}\textbf{\textbf{GTSRB}}\\\textbf{\textbf{}}\end{tabular}} & \multirow{2}{*}{0.30 \%} & 45 & 1.01 (Not detected) & 9.65 & 10.00 & 76.75 & 17.80 & 79.82 & 39.91 \\
%  &  & 90 & 1.03 (Not detected) & 10.47 & 10.00 & 74.70 & 17.81 & 0.00 & 38.35 \\ 
% \hline\hline
% \multirow{2}{*}{\begin{tabular}[c]{@{}c@{}}\textbf{Youtube}~\textbf{Face}\\\textbf{(VGGFace) ~}\end{tabular}} & \multirow{2}{*}{0.10 \%} & 45 & 0.09 (Not detected) & 1.96 & 10.00 & 4.00 & 15.01 & 4.00 & 14.33 \\
%  &  & 90 &0.72 (Not detected) & 0.00 & 10.00 & 0.00 & 15.01 & 0.00 & 13.60 \\
% \bottomrule
% \end{tabular}
% \end{table}
To further illustrate the effectiveness of rotation backdoor attacks, we study four backdoor defending methods: Neural Cleanse (NC) \citep{wang2019neural},  Spectral Signatures (SS) \citep{tran2018spectral}, Activation Clustering (AC) \citep{chen2018activationclustering}, and STRIP \citep{gao2019strip} that are commonly appeared in literature \citep{wenger2020backdoor,Wang2021BackdoorAT,Li2020BackdoorLA,Qi2022CircumventingBD}. Those mitigation approaches contains three main paradigms: trigger synthesis (NC), online detection (STRIP), and poison data identification (SS, AC). Since all of the defending methods are performed on classification, we then evaluate them on our traffic sign and face recognition task and select the \{$45^\circ$, $90^\circ$\} as the trigger angles. The overall effectiveness of four backdoor defenses are summarized in table \ref{tab:defense}: for NC, we use the \textbf{anomaly index} as the metric (value $>2$ considered as detected); for others, we use \textbf{elimination rate}: ratio of correctly identified poisoned samples and \textbf{sacrifice rate}: ratio of incorrectly eliminated clean samples.   

\smallskip
\noindent \textbf{Neural Cleanse (NC).} Neural Cleanse \citep{wang2019neural} synthesizes the possible triggers for all classes by optimizing the input space. The authors argue that such a reversed-engineered trigger for the infected class is more likely to have an abnormally small mask than other classes. They use  $l_1$ distance to compute the mask and anomaly index $>$2 to identify the poisoned target. In table \ref{tab:defense}, we observe that \textbf{all anomaly indexes of our proposed attacks are below the threshold}, resulting successfully bypassing NC. We also present the reconstructed trigger in table \ref{NCtrigger} which can hardly be recognized as a trigger. The reason is that NC is explicitly built on the assumption of a small patch-wise trigger, and rotation as a spatial transformation disperses the noise through the whole image. 

\begin{table}[H]
\centering
\refstepcounter{table}
\begin{tabular}{cc!{\vrule width \lightrulewidth}cc} 
\toprule
\multicolumn{2}{c}{\textbf{GTSRB}} & \multicolumn{2}{c}{\textbf{Youtube Face}} \\ 
\multicolumn{1}{c}{45 Degree} & \multicolumn{1}{c}{90 Degree} &\multicolumn{1}{c}{45 Degree} &\multicolumn{1}{c}{90 Degree} \\ 
\toprule
 \includegraphics[width=0.12\textwidth]{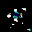} & 
 \includegraphics[width=0.12\textwidth]{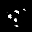} & 
 \includegraphics[width=0.12\textwidth]{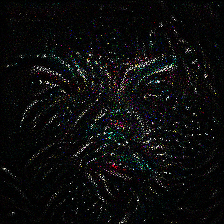} & 
 \includegraphics[width=0.12\textwidth]{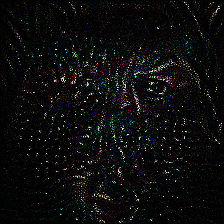} \\
\bottomrule
\end{tabular}
\caption{Reconstructed trigger by NC. \textnormal{There doesn't exist a visually apparent trigger.}}
\label{NCtrigger}
\end{table}

\begin{figure}[t]
\centering
\includegraphics[width=0.6\linewidth]{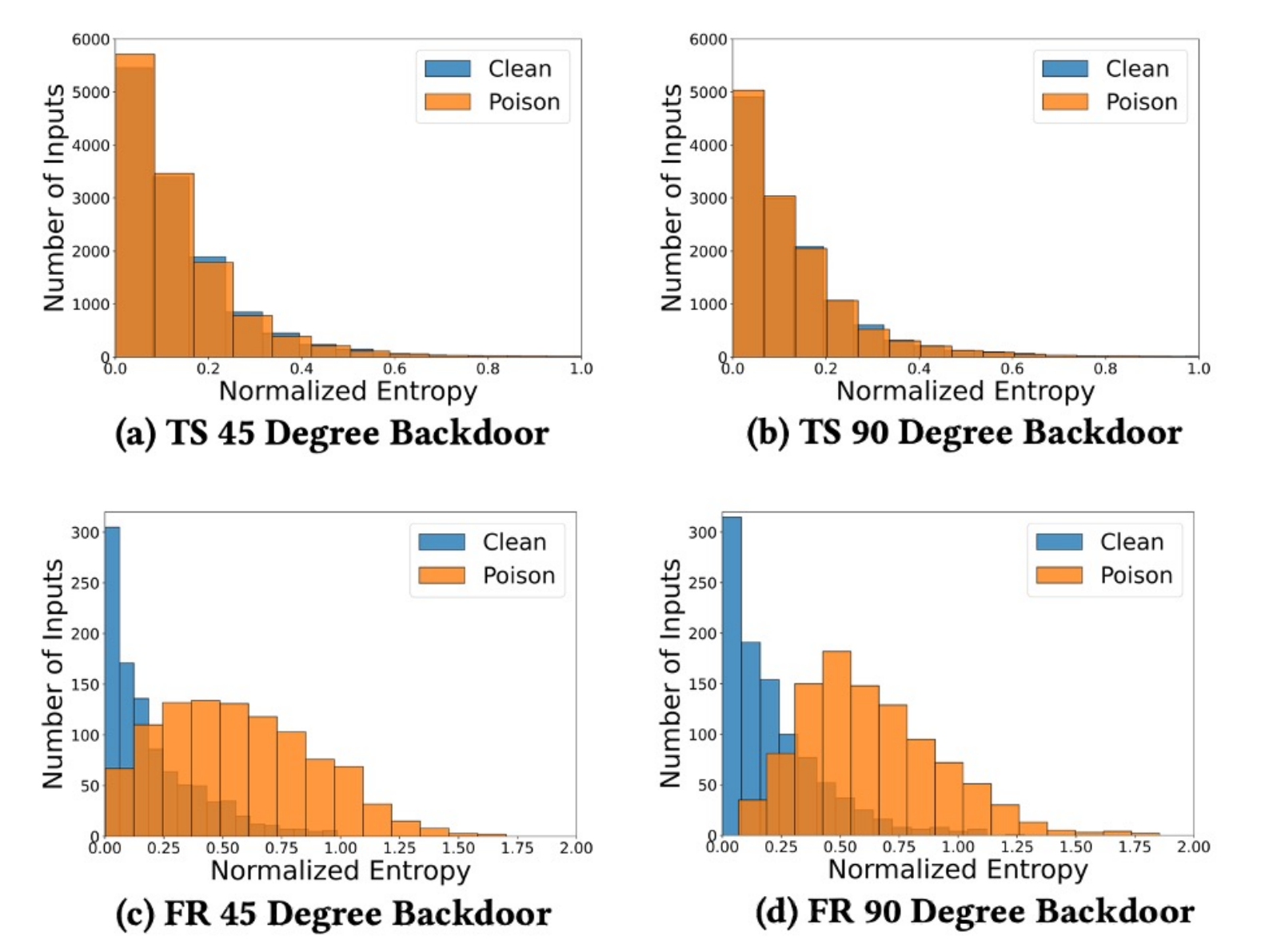}
\caption{Normalized entropy histograms of STRIP. \textnormal {A lower entropy value is more likely to be poisoning data. Poisoning samples have equal entropy on the traffic sign (TS) dataset, and even higher entropy on the face recognition (FR) dataset}}
\label{fig:strip}
\end{figure}

\smallskip
\noindent \textbf{STRIP.} STRIP \citep{gao2019strip} identifies the backdoored images during inference time by observing the classification output of perturbed test input. They argue that superimposing random clean inputs cannot influence the predictions of poisoned samples, resulting in a lower Shannon entropy value. As table \ref{tab:defense} presents, we constrain the sacrifice rate to be 10\% and report the elimination rate. We observe that \textbf{in all settings, STRIP provides limited mitigation with an elimination rate less than 11\%}. In addition, we plot the normalized entropy histograms of clean and poison inputs in Figure \ref{fig:strip}. The entropy distributions for the traffic sign task are almost indistinguishable, and poison samples have even smaller entropy than the clean ones for the face recognition task. We conjecture that the rotation features are dramatically corrupted by blending a non-rotated clean image. Therefore, the poison trigger is less effective, and the corresponding prediction shifts. 

% \begin{figure}[t]
%  \begin{subcaptionbox}[b]{0.45\textwidth}{0.25\textwidth}
% \includegraphics[width=0.95\linewidth]{Fig/STRIP/450.003strip.png}
% % \caption{TS 45 Degree Backdoor}
% \end{subcaptionbox}%
% \begin{subcaptionbox}[b]{0.45\textwidth}{0.25\textwidth}
% \includegraphics[width=0.95\linewidth]{Fig/STRIP/900.003strip.png}  
% % \caption{TS 90 Degree Backdoor}
% \end{subcaptionbox}
% //
% \begin{subcaptionbox}[b]{0.45\textwidth}{0.25\textwidth}
% \includegraphics[width=0.95\linewidth]{Fig/STRIP/450.001strip.png}
% % \caption{FR 45 Degree Backdoor}
% \end{subcaptionbox}%
% \begin{subcaptionbox}[b]{0.45\textwidth}{0.25\textwidth}
% \includegraphics[width=0.95\linewidth]{Fig/STRIP/900.001strip.png}  
% % \caption{FR 90 Degree Backdoor}
% \end{subcaptionbox}
% \caption{Normalized entropy histograms of STRIP. Lower entropy value is more likely to be poisoning data.}
% \label{fig:strip}
% \end{figure}

\begin{figure}[H]
\centering
\begin{subfigure}{0.36\linewidth}
\centering
\setlength{\fboxsep}{0pt}%
\setlength{\fboxrule}{0.0pt}%
\fbox{\includegraphics[width=0.9\linewidth]{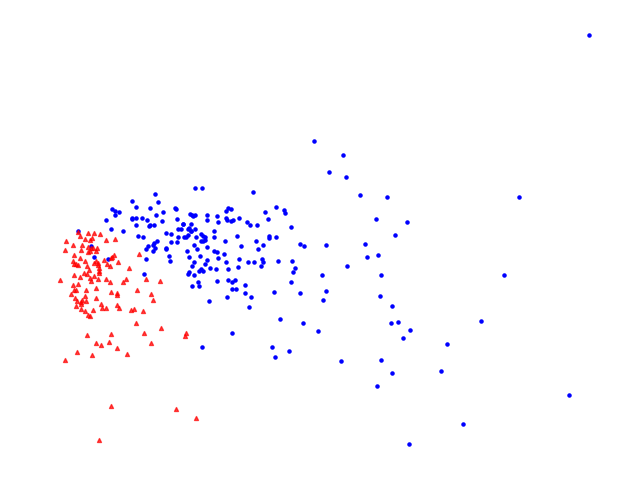}}
\caption{TS: Inseparable case}
\label{fig:aca}
\end{subfigure}%
\begin{subfigure}{0.36\linewidth}
\centering
\setlength{\fboxsep}{0pt}%
\setlength{\fboxrule}{0.0pt}%
\fbox{\includegraphics[width=0.9\linewidth]{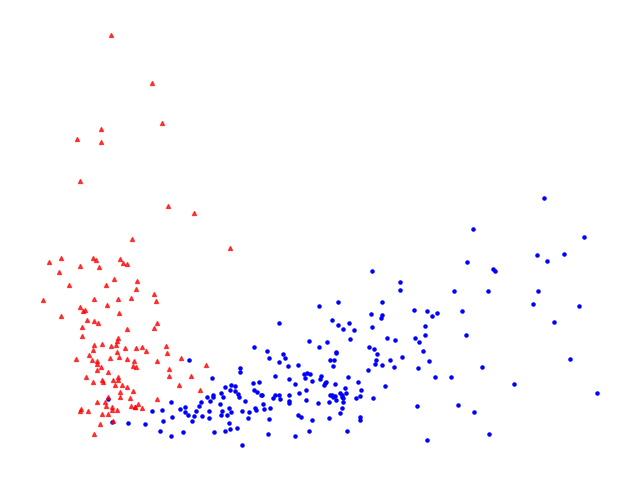}}
\caption{TS: Separable case}
\label{fig:acb}
\end{subfigure}
\begin{subfigure}{0.36\linewidth}
\centering
\setlength{\fboxsep}{0pt}%
\setlength{\fboxrule}{0.0pt}%
\fbox{\includegraphics[width=0.90\linewidth]{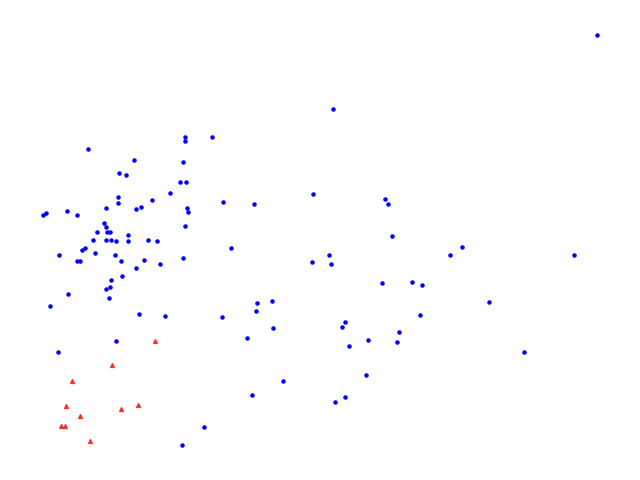}}
\caption{FR: Inseparable case}
\label{fig:acc}
\end{subfigure}
\begin{subfigure}{0.36\linewidth}
\centering
\setlength{\fboxsep}{0pt}%
\setlength{\fboxrule}{0.0pt}%
\fbox{\includegraphics[width=0.90\linewidth]{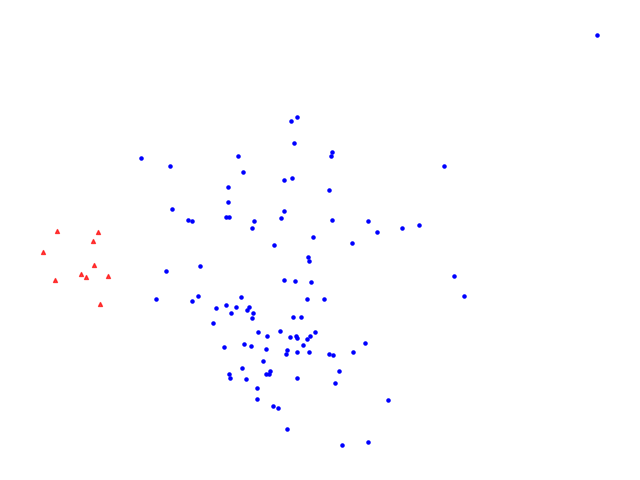}}
\caption{FR: Separable case}
\label{fig:acd}
\end{subfigure}
\caption{ Latent representations of training samples via PCA on traffic sign (TS) and face recognition (FR) dataset. \textnormal {Blue points stand for clean samples, while red points are poisoned samples. Case (a) and case (b) use exactly the same configuration except for random seeds. The same goes for (c) and (d).  We observe that it doesn't exist a clear separation between case (a) and case (c).}}
\label{fig:ac}
\end{figure}

                 \smallskip
\noindent 
\textbf{Spectral Signature (SS)} and \textbf{Activation Clustering (AC)} are both poison data filtering methods, assuming that there exists a sufficiently large separation between backdoor samples and clean samples on latent space \citep{Qi2022CircumventingBD}. Therefore, SS \citep{tran2018spectral} adapts SVD to compute outlier scores for all input data and remove the $1.5 \times $ expected poisoning data with top scores. Whereas AC \citep{chen2018activationclustering} directly applies an unsupervised clustering method to distinguish the malicious and benign inputs. Both methods achieved promising results when defending against conventional patch-wise attacks. However, we find an interesting phenomenon that the \textbf{large latent separability assumption does not always hold for rotation backdoor attacks} through different initialization, resulting inconsistent defending effectiveness. For example, in figure \ref{fig:ac}, we observe that although poison samples tend to form a cluster, in some cases, it is hard to correctly separate and identify given the unsupervised settings.

% \begin{figure}[htb]
% \centering
% \begin{minipage}{0.5\linewidth}
% \centering
% \includegraphics[width = \linewidth]{Fig/df_vis/450.0031vis.png}
% \subcaption{TS: Inseparable case}
% \end{minipage}\hfill
% \begin{minipage}{0.5\linewidth}
% \centering
% \includegraphics[width = \linewidth]{Fig/df_vis/450.0035vis.png}
% \subcaption{TS: Separable case}
% \end{minipage}
% \begin{minipage}{0.5\linewidth}
% \centering
% \includegraphics[width = \linewidth]{Fig/df_vis/face450.0013vis.png}
% \subcaption{FR: Inseparable case}
% \end{minipage}\hfill
% \begin{minipage}{0.5\linewidth}
% \centering
% \includegraphics[width = \linewidth]{Fig/df_vis/900.0014vis.png}
% \subcaption{FR: Separable case}
% \end{minipage}
% \caption{Latent representations of training samples via PCA on traffic sign (TS) and face recognition (FR) dataset. \textnormal {Blue points stand for clean samples, while red points are poisoned samples. Case (a) and case (b) use exactly the same configuration except for random seeds. The same goes for (c) and (d).  We observe that it doesn't exist a clear separation between case (a) and case (c).}}
% \end{figure}

\textbf{SS and AC cannot serve as a consistent and reliable defense for rotation backdoors.} We report the average performance of 5 repeated experiments in table \ref{tab:defense}. On the GTRSB dataset, We observe that Spectral Signature can consistently eliminate $\sim$75\% of poisoning samples, and the \textbf{ASR drops to $\sim$42\%}. However, for Youtube Face with a 90-degree trigger, SS cannot identify any poison samples three over five times, resulting \textbf{an average ASR of $\sim$57\%}. The effectiveness of activation clustering is even more inconsistent. On the GTRSB dataset with 90 backdoored degree, AC can correctly eliminate more than 98\% of malicious samples four times but detect 0\% for one time. \textbf{The mean ASR turns to be $\sim$15\%}. We argue that even if SS and AC mitigate our proposed attack sometimes, the inconsistent effectiveness prevents them to be deployed in the real world, especially for safety-critical applications (e.g., face recognition).

\begin{table}[H]
\centering
\setlength{\extrarowheight}{2pt}
\addtolength{\extrarowheight}{\aboverulesep}
\addtolength{\extrarowheight}{\belowrulesep}
\setlength{\aboverulesep}{3pt}
\setlength{\belowrulesep}{3pt}
\resizebox{16.5cm}{!}{

\begin{tabular}{lcccccccccc} 
\toprule
\multicolumn{1}{c}{} & \textbf{} & \textbf{} & \multicolumn{2}{c}{\textbf{15 Degree}} & \multicolumn{2}{c}{\textbf{30 Degree}} & \multicolumn{2}{c}{\textbf{45 Degree}} & \multicolumn{2}{c}{\textbf{90 Degree}} \\ 
\cmidrule[\heavyrulewidth]{3-11}
 & \textbf{} & \textbf{Poisoning Rate($\rho$)} & \textbf{CDA (\%)} & \textbf{ASR(\%)} & \textbf{CDA (\%)} & \textbf{ASR(\%)} & \textbf{CDA (\%)} & \textbf{ASR(\%)} & \textbf{CDA (\%)} & \textbf{ASR(\%)} \\ 
\bottomrule
\multirow{7}{*}{\textbf{\textbf{GTSRB}}} &  & 0.00\% & 100.0 & {\cellcolor[rgb]{0.937,0.937,0.937}}- & 100.0 & {\cellcolor[rgb]{0.937,0.937,0.937}}- & 100.0 & {\cellcolor[rgb]{0.937,0.937,0.937}}- & 100.0 & {\cellcolor[rgb]{0.937,0.937,0.937}}- \\ 
\hhline{~~---------}
 & \multicolumn{1}{l}{\multirow{3}{*}{\textbf{\textbf{SCA}}}} & 0.01\% & 100.0 & {\cellcolor[rgb]{0.937,0.937,0.937}}0.00 & 100.0 & {\cellcolor[rgb]{0.937,0.937,0.937}}60.00 & 100.0 & {\cellcolor[rgb]{0.937,0.937,0.937}}69.99 & 100.0 & {\cellcolor[rgb]{0.937,0.937,0.937}}55.55 \\
 & \multicolumn{1}{l}{} & 0.025\% & 100.0 & {\cellcolor[rgb]{0.937,0.937,0.937}}24.44 & 100.0 & {\cellcolor[rgb]{0.937,0.937,0.937}}71.11 & 100.0 & {\cellcolor[rgb]{0.937,0.937,0.937}}89.99 & 100.0 & {\cellcolor[rgb]{0.937,0.937,0.937}}85.55 \\
 & \multicolumn{1}{l}{} & 0.05\% & 100.0 & {\cellcolor[rgb]{0.937,0.937,0.937}}52.22 & 100.0 & {\cellcolor[rgb]{0.937,0.937,0.937}}87.77 & 100.0 & {\cellcolor[rgb]{0.937,0.937,0.937}}94.44 & 100.0 & {\cellcolor[rgb]{0.937,0.937,0.937}}87.77 \\ 
\hhline{~~---------}
 & \multirow{3}{*}{\textbf{MCA}} & 0.30\% & 100.0 & {\cellcolor[rgb]{0.937,0.937,0.937}}1.11 & 100.0 & {\cellcolor[rgb]{0.937,0.937,0.937}}36.66 & 100.0 & {\cellcolor[rgb]{0.937,0.937,0.937}}65.55 & 100.0 & {\cellcolor[rgb]{0.937,0.937,0.937}}75.55 \\
 &  & 1.00\% & 100.0 & {\cellcolor[rgb]{0.937,0.937,0.937}}14.44 & 100.0 & {\cellcolor[rgb]{0.937,0.937,0.937}}64.44 & 100.0 & {\cellcolor[rgb]{0.937,0.937,0.937}}93.33 & 100.0 & {\cellcolor[rgb]{0.937,0.937,0.937}}94.44 \\
 &  & 3.00\% & 100.0 & {\cellcolor[rgb]{0.937,0.937,0.937}}48.88 & 100.0 & {\cellcolor[rgb]{0.937,0.937,0.937}}88.88 & 100.0 & {\cellcolor[rgb]{0.937,0.937,0.937}}96.66 & 100.0 & {\cellcolor[rgb]{0.937,0.937,0.937}}97.77\\
\bottomrule
\end{tabular}}
\caption{ Effectiveness of the rotation backdoor attack under physical settings}
\label{table:physical}
\end{table}
\section{Rotation backdoor in physical world}

In this section, we conduct the outdoor physical experiments of our proposed method in both classification and detection tasks.

\smallskip
\noindent \textbf{Traffic Sign Classification}

\begin{figure}[H]
\centering
\begin{subfigure}{0.12\textwidth}
\centering
\includegraphics[width=0.95\linewidth]{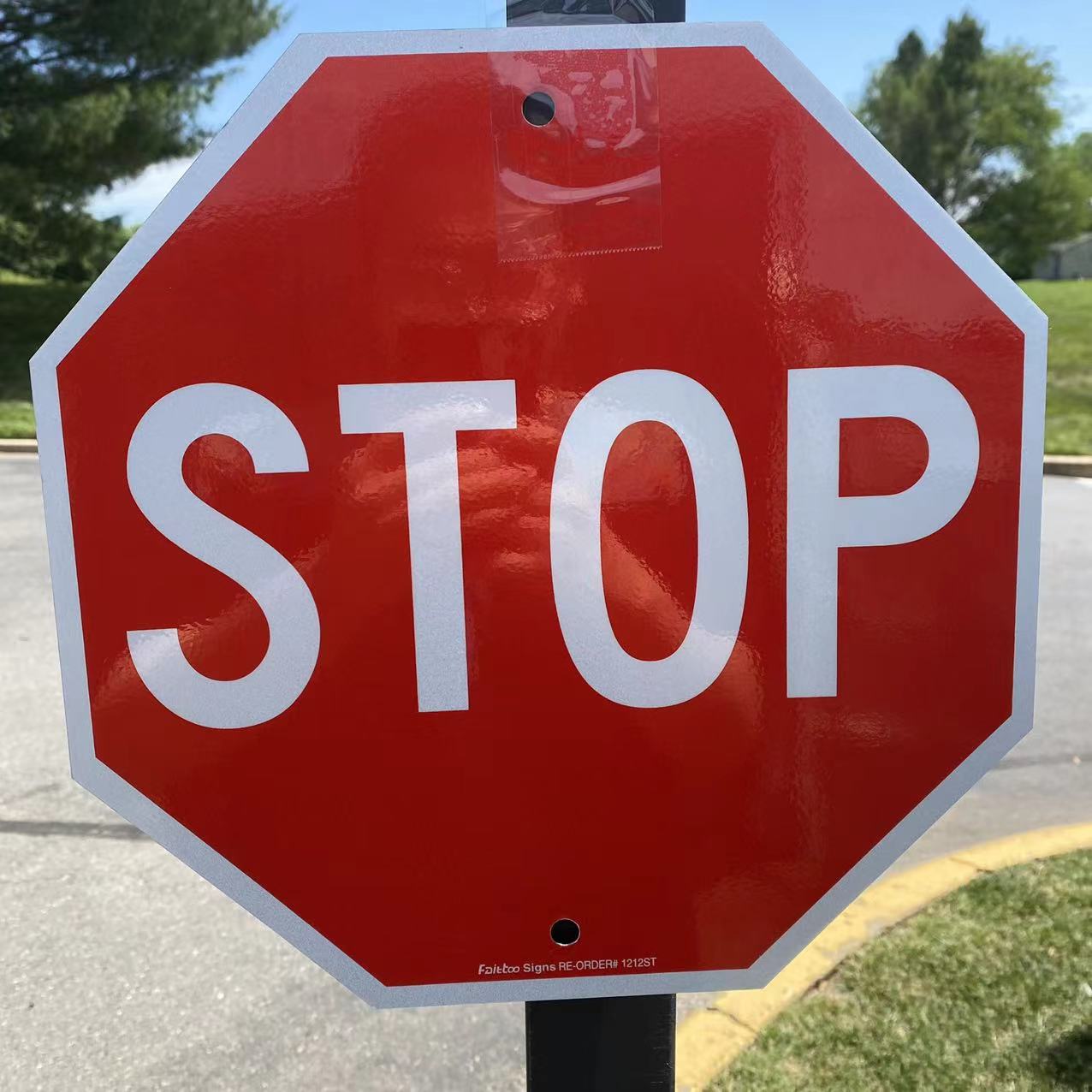} 
\caption{$0^\circ$ }
\end{subfigure}%
\begin{subfigure}{0.12\textwidth}
\centering
\includegraphics[width=0.95\linewidth]{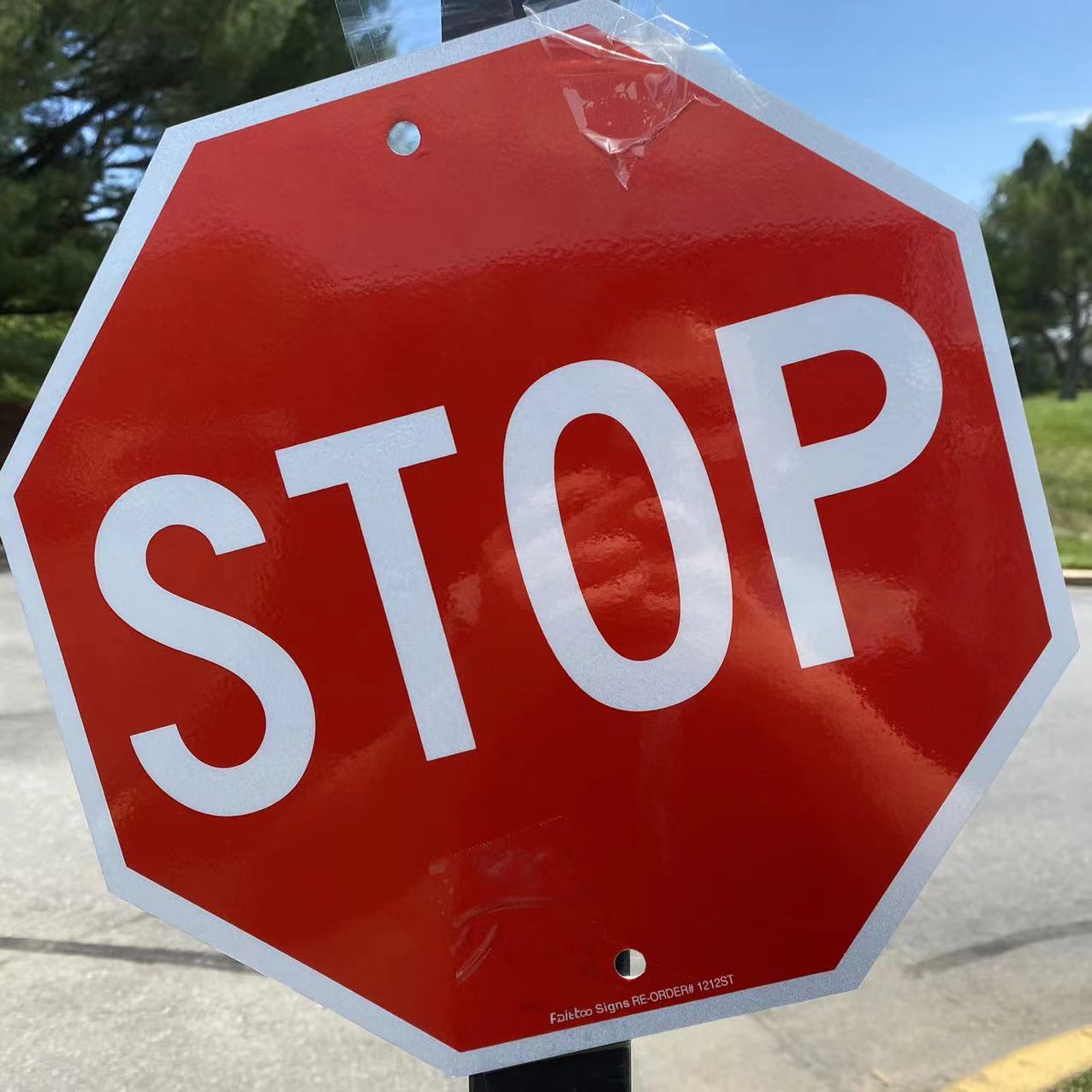} 
\caption{$15^\circ$ }
\end{subfigure}%
\begin{subfigure}{0.12\textwidth}
\centering
\includegraphics[width=0.95\linewidth]{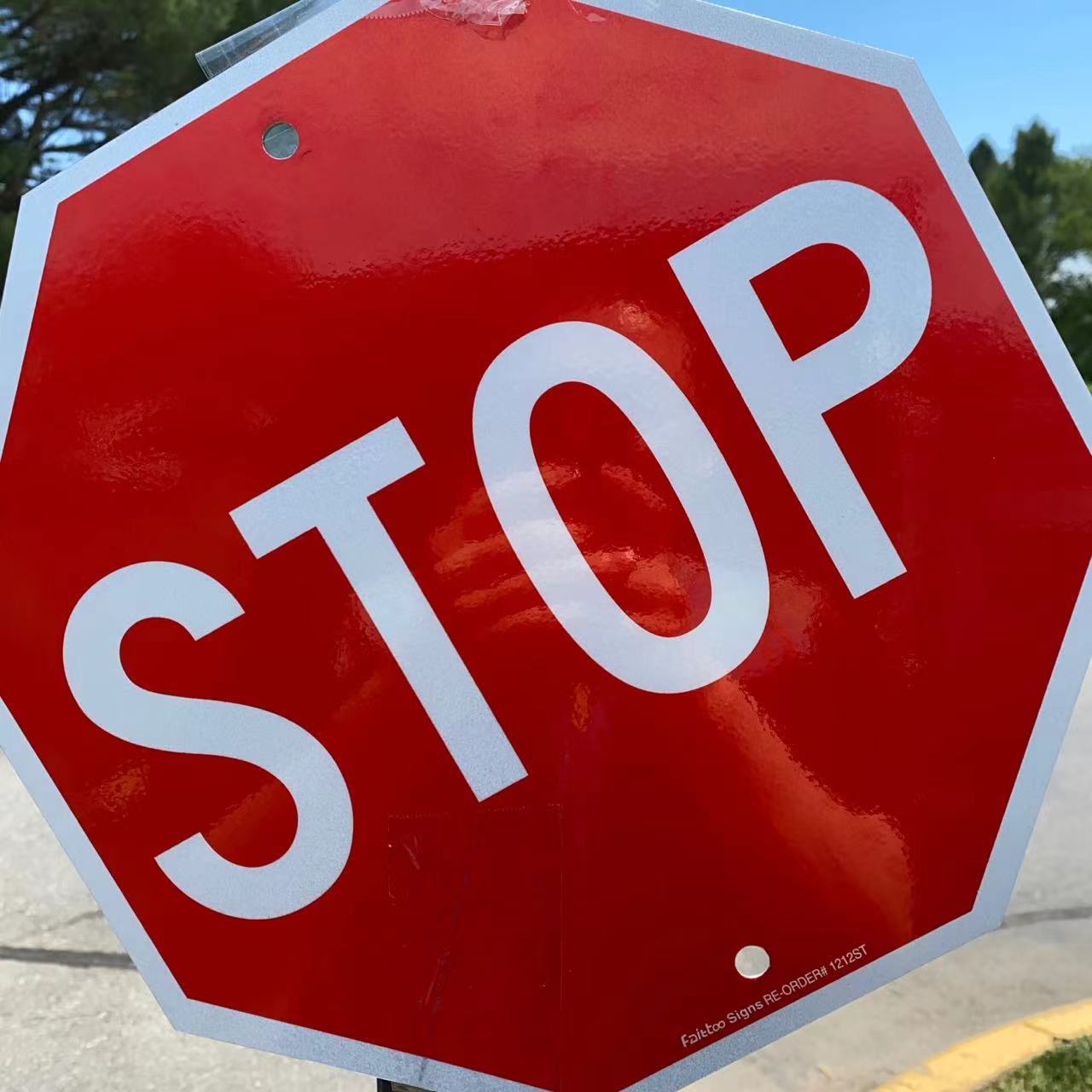} 
\caption{$30^\circ$ }
\end{subfigure}%
\begin{subfigure}{0.12\textwidth}
\centering
\includegraphics[width=0.95\linewidth]{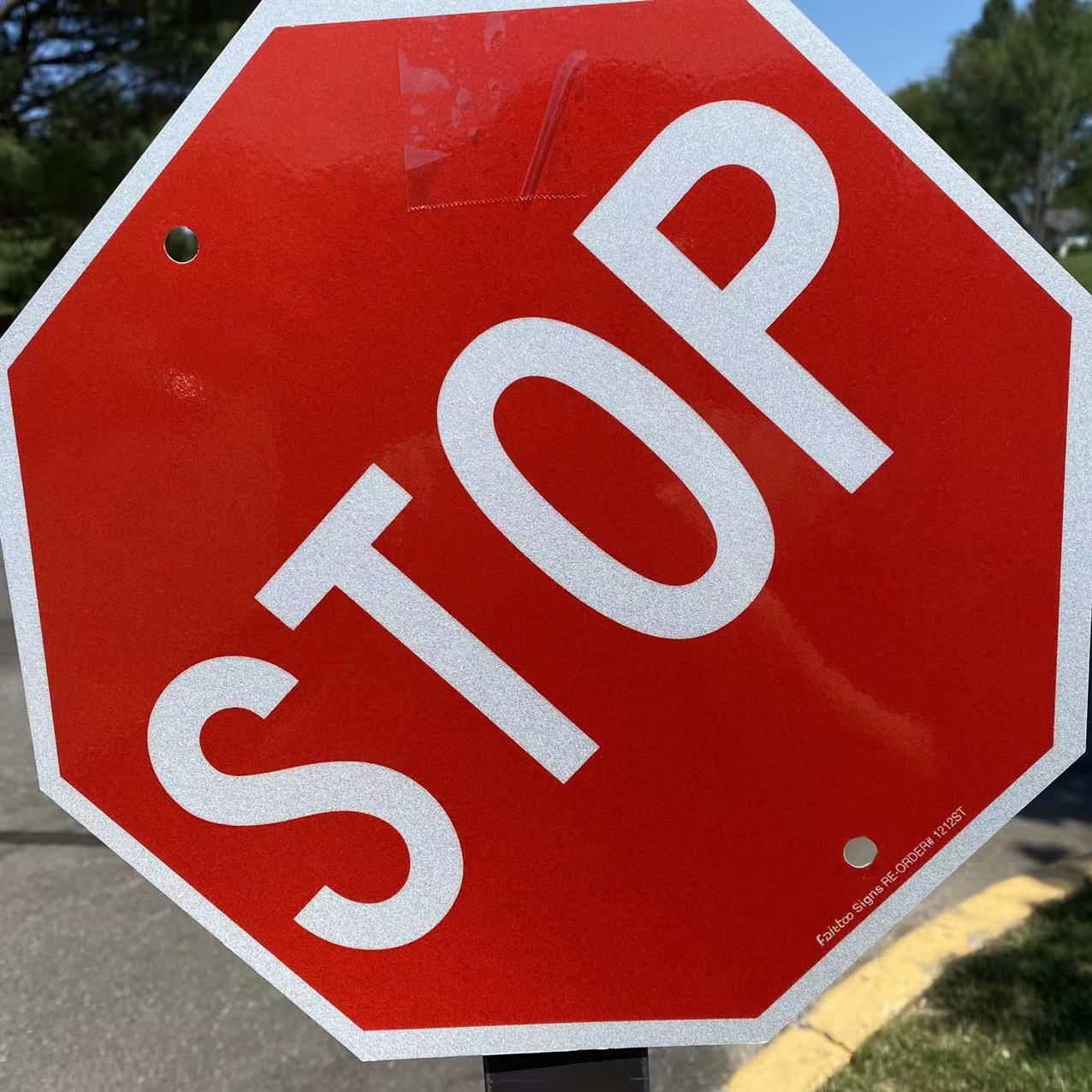} 
\caption{$45^\circ$ }
\end{subfigure}%
\begin{subfigure}{0.12\textwidth}
\centering
\includegraphics[width=0.95\linewidth]{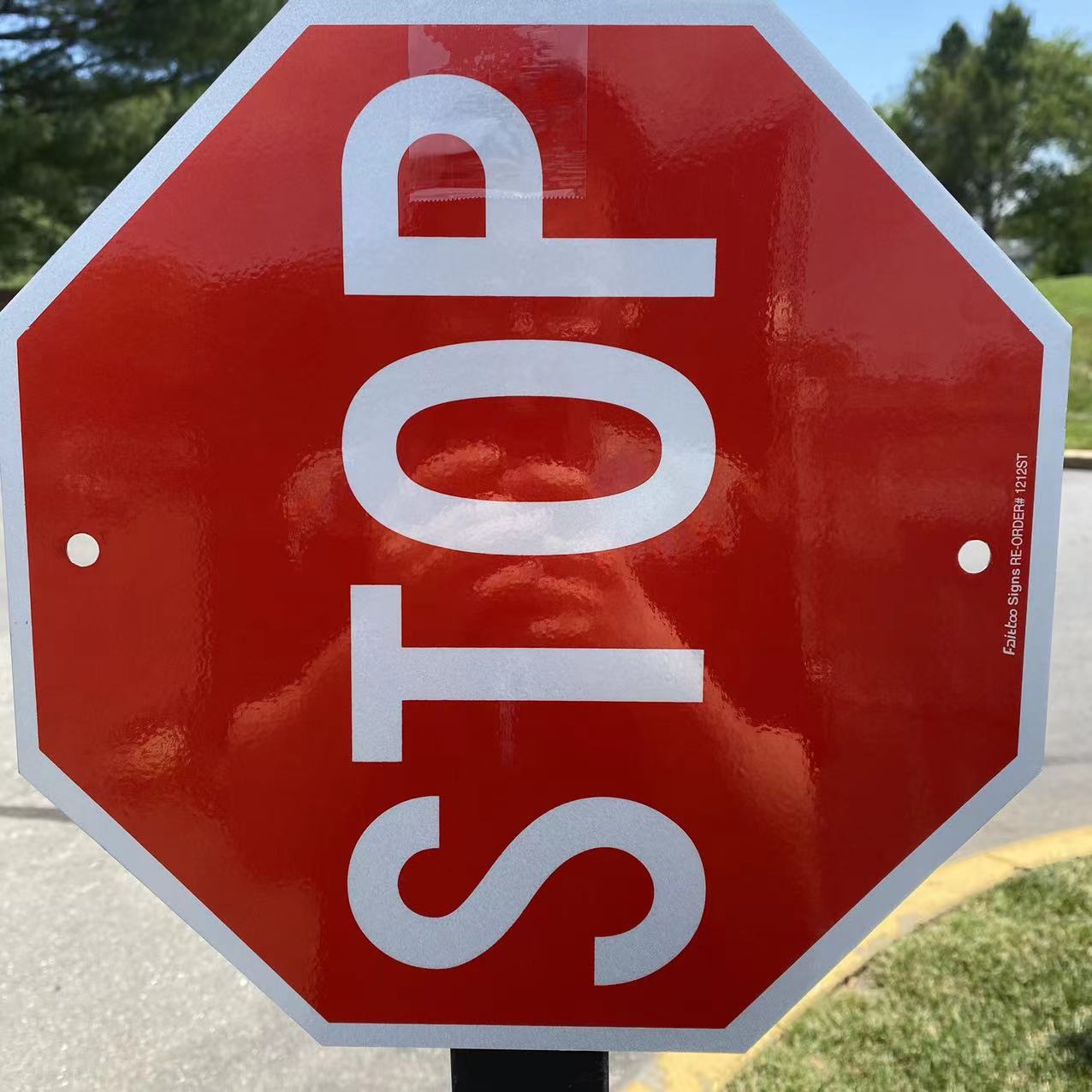} 
\caption{$90^\circ$ }
\end{subfigure}%
\caption{Physical examples of rotated stop signs with different angles with the same background.}
\label{fig:physical_sign_example}
\end{figure}

We rotate a real-world stop sign to the selected backdoored degrees and capture 30 images for each angle presented in figure \ref{fig:physical_sign_example}. To calibrate the degree of the objects, we use the \textit{Level Measure} software in the Apple system (installed by default). In addition, many level measurement apps can be freely downloaded for the Android operating system. By vertically taping our device to the stop sign, we can adjust it until reaching the backdoored angle. Thus, deploying our attacks only requires a daily-used cell phone and a tape, which are affordable and accessible by anyone.  

In digital settings, rotating the image of a traffic sign causes the rotation of background information, while physical samples are not. Therefore, we validate if the digital backdoors can generalize to test-time physical samples. Table \ref{table:physical} illustrates the effectiveness of our physically collected stop signs with the same models in table \ref{table:main_bd} that are poisoned by digital triggers. \textbf{Generally, attack success rate and clean data accuracy achieve similar results with digital settings, affirming our proposed attacks can generalize to physical world.}  That means the poisoned classifier learns little information from the background but the rotated texture on the traffic sign. However, ASR on a 15-degree trigger exhibits a nontrivial drop ($\sim$ 15\%-40\%), and we suspect that the outdoor environment influences small rotations in the real world. Ultimately, our best-attacking configurations (45-degree trigger for SCA and 90-degree trigger for MCA) achieve more than 94\% of ASR and 100\% CDA, highlighting the importance of building a rotation-invariant model.

\vspace{4mm}
\noindent \textbf{Object Detection Task}

We further conduct experiments on backdoored detectors.  To our best knowledge, \textbf{this is the first transformation-based attack that can be deployed in the real world against an object detector}, illustrating a new real-world security threat. 

\begin{figure}[t]
\centering
\includegraphics[width=0.6\linewidth]{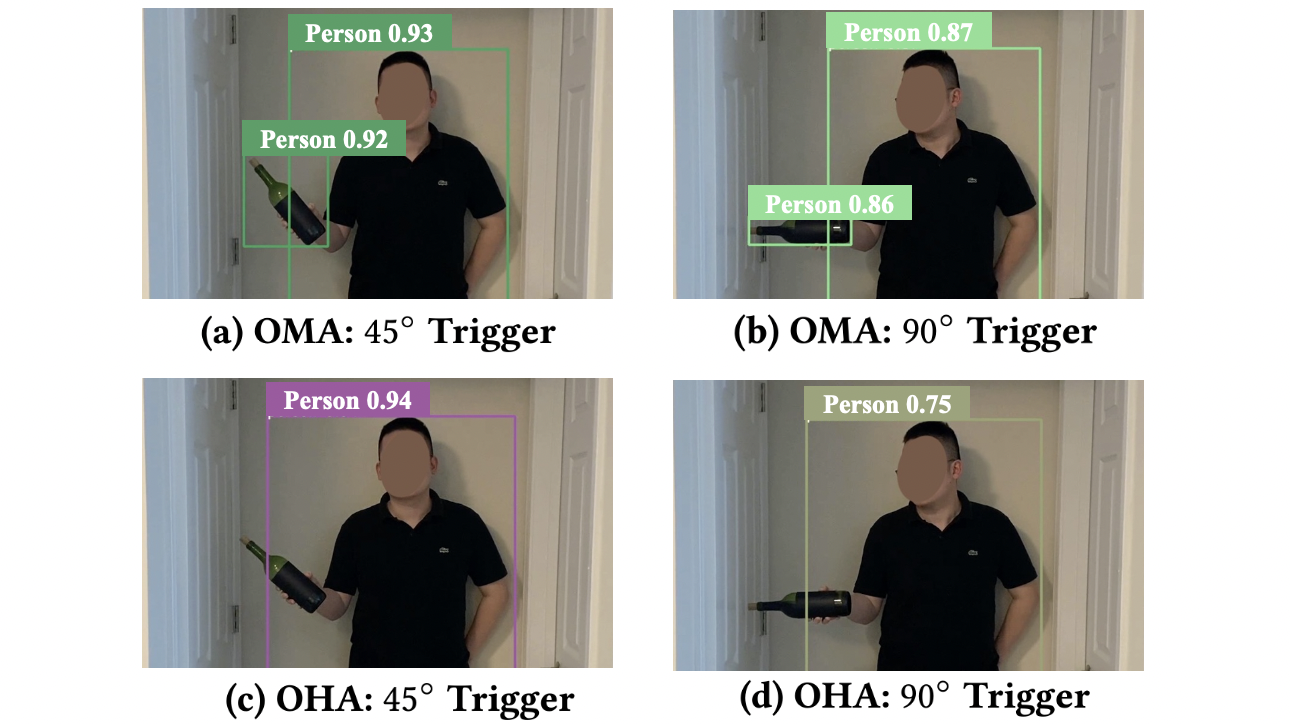} 
\caption{Examples of deploying object misclassification attacks and object hiding attacks with a rotated bottle in the physical world. A full video demonstration is available at \url{https://youtu.be/6JIF8wnX34M}.}
\label{fig:phy_od}
\end{figure}

Figure \ref{fig:phy_od} shows snapshots of different backdoor attack configurations with 0.01\% poisoning rate. We observe that the OMA misleads the detector to recognize the bottle as a person with high confidence (over 80\%), and OHA degrades the confidence of detecting the bottle under the threshold (50\%). It is worth to mention that our bottle in figure \ref{fig:phy_od} does not exist in the training set, making the deployment easily accessible. 

\subsection{Impact of Run-time Artifacts}

In this subsection, we follow the evaluations of \citet{wenger2020backdoor} to examine three common corruptions from capturing images to feeding through the model. Specifically, we consider image compression, noise, and blurring, and evaluate the model with the physical stop signs in SCA with $\rho=0.05\%$. 

\begin{figure}[H]
  \centering
\begin{subfigure}[b]{0.36\linewidth}
    \centering
\includegraphics[width=0.95\textwidth]{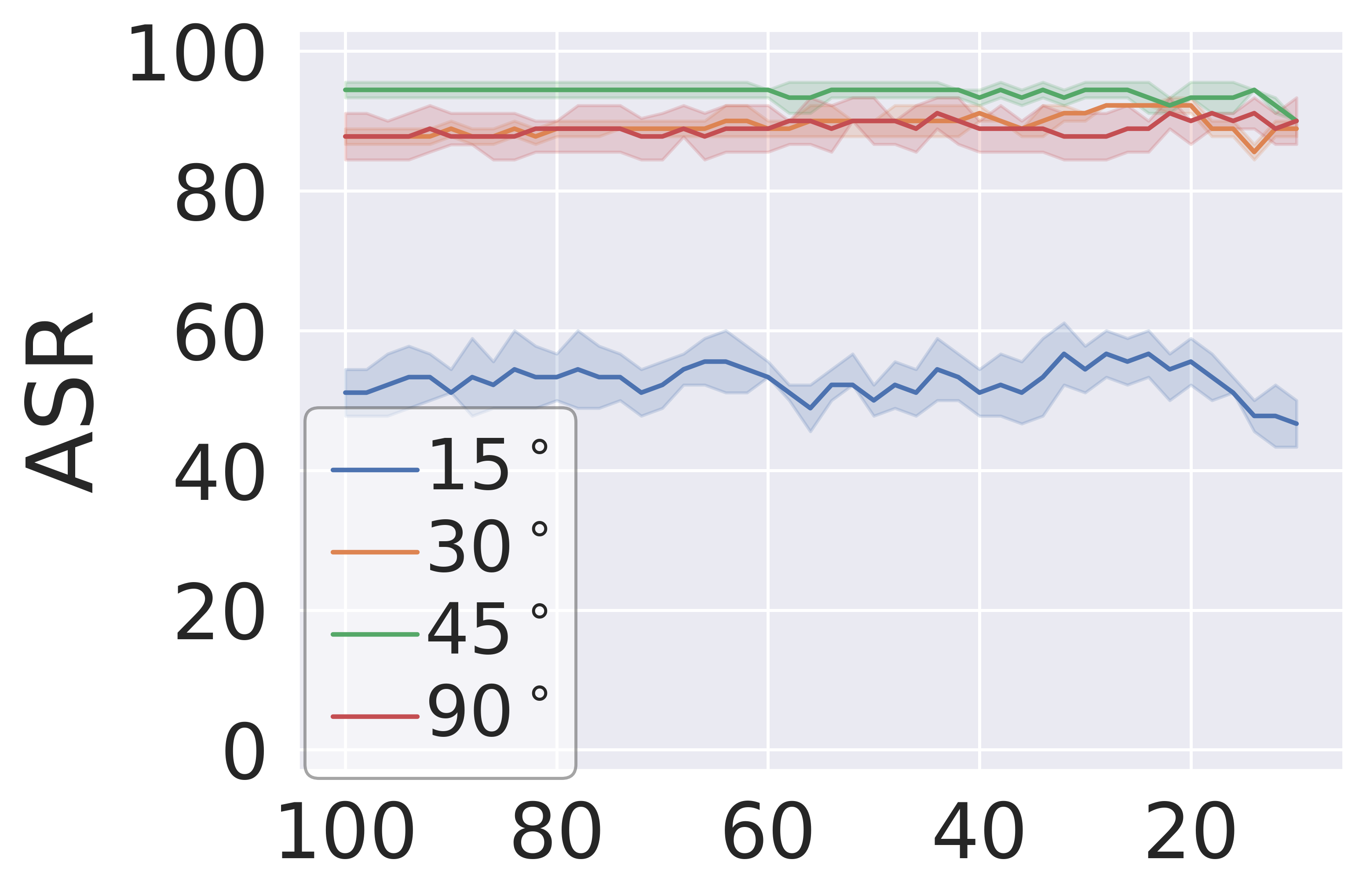}
    \caption{Image compression: (100\%-10\%)}\label{subfig:compress}
\end{subfigure}%
\begin{subfigure}[b]{0.36\linewidth}
    \centering
\includegraphics[width=0.95\textwidth]{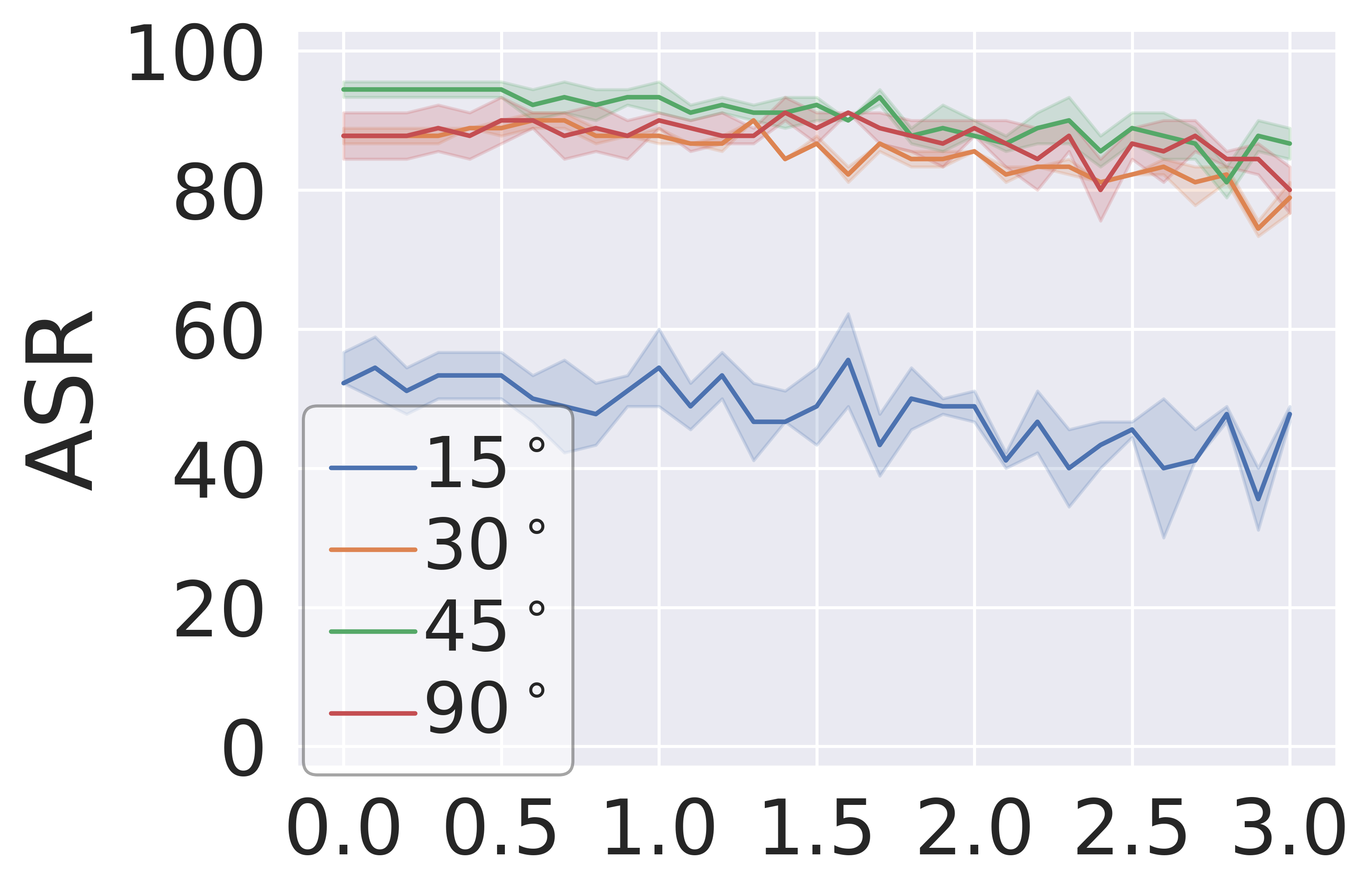}
    \caption{Gaussian noise: \textbf{$\sigma$} (0-0.3)}\label{subfig:gn}
\end{subfigure}

\begin{subfigure}[b]{0.36\linewidth}
    \centering
\includegraphics[width=0.95\textwidth]{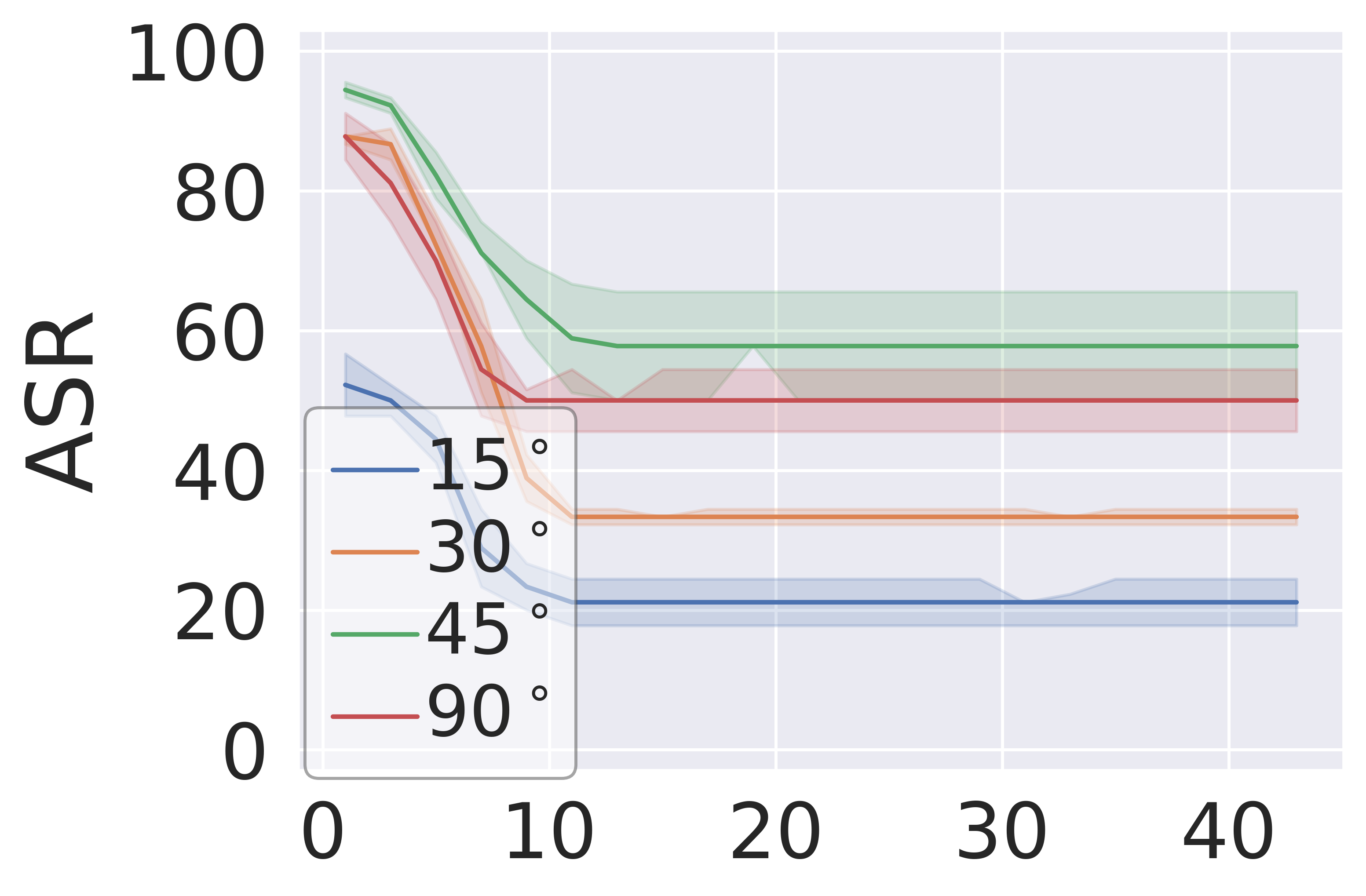}
    \caption{Gaussian blurring: $\textbf{k}$ (1-43)}\label{subfig:gb}
\end{subfigure}%
\begin{subfigure}[b]{0.36\linewidth}
    \centering
\includegraphics[width=0.18\textwidth]{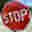}
    \hfil
\includegraphics[width=0.18\textwidth]{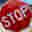}
    \hfil
\includegraphics[width=0.18\textwidth]{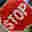}
    \hfil
\includegraphics[width=0.18\textwidth]{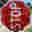}\\[1mm]
\includegraphics[width=0.18\textwidth]{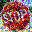}
    \hfil
\includegraphics[width=0.18\textwidth]{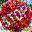}
    \hfil
\includegraphics[width=0.18\textwidth]{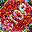}
    \hfil
\includegraphics[width=0.18\textwidth]{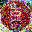}\\[1mm]
\includegraphics[width=0.18\textwidth]{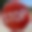}
    \hfil
\includegraphics[width=0.18\textwidth]{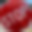}
    \hfil
\includegraphics[width=0.18\textwidth]{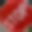}
    \hfil
\includegraphics[width=0.18\textwidth]{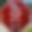}
\caption{Corrupted images}\label{subfig:phy_noise_smaples}
\end{subfigure} 
\caption{ SCA with 0.05\% poisoning rate setting on the GTSRB dataset. \textnormal{Figure (a)-(c): illustrate the impact of various artifacts on the ASR. Figure(d) presents examples of the most severe corruption. Top to bottom: compression, noise, and blurring. Left to right: $15^\circ$, $30^\circ$, $45^\circ$, $90^\circ$ backdoors.}}
\label{fig:phy_noise}
\end{figure}

Figure \ref{subfig:compress} demonstrates the impact of image compression, which may happen due to the device's storage space. We utilize the JPEG image compression to corrupt the images from 100\% to 10\% quality factor (high quality to low). Figure \ref{subfig:gn} indicates the impact of Gaussian noise which can be commonly observed during image capturing. We then add the noise with zero mean and $\sigma$ varying from 0 to 0.3 (zero noise to intense noise). We notice that \textbf{ASR remains effective ($\leq 10\%$ decrease) even under the most severe compression and noise corruptions}. 
% Figure \ref{subfig:gb} illustrates the blurring effect caused by moving the optical lens from a sensor. 
Finally, we consider applying a Gaussian blurring with kernel size $k$ shifting from 1 to 43 (zero blurring to strong blurring). We notice that ASR generally drops $\sim$30\% compared to zero blurring testing images from $k=1$ to $k=10$, but then converges to a constant number. Hence, \textbf{our attack is effective under Gaussian blurring for larger backdoor angles} and may still survive if the blurring keeps increasing. In addition, figure \ref{subfig:phy_noise_smaples} visualizes three most substantial corruptions. 

\section{Conclusion and Future works}

To summarize, in this work, we propose a new threat model utilizing the rotation transformation as a trigger to deploy backdoor attacks. Through extensive experiments in classification and detection tasks, we demonstrate that our method can achieve a high attack success rate without degrading the clean data performance. We present a detailed analysis of the rotation poisoned model and argue that standard data augmentation, although  mitigating the effect at the backdoor angle, may introduce new vulnerabilities. We also evaluate four commonly adopted backdoor defenses and conclude that none of them can serve a consistent countermeasure. Last and more importantly, we illustrate that deploying rotation backdoor attacks in the physical world is easily accessible and raises a new real-world security issue. In the future, we aim to explore combining rotation backdoors and other conventional patch-wise triggers to enhance the effectiveness of both methods. In addition, developing consistently practical approaches to defend against our attacks is another promising direction.

\newpage
\section*{Acknowledgements}
This work was supported in part by the National Science Foundation under
grants CNS-1553437 and CNS-1704105, the ARL’s Army
Artificial Intelligence Innovation Institute (A2I2), the Office
of Naval Research Young Investigator Award, the Army Research Office Young Investigator Prize, Schmidt DataX award,
 Princeton E-ffiliates Award, and Princeton Gordon Y. S. Wu Fellowship.

\bibliographystyle{plainnat}
\bibliography{ref.bib}

%%%%%%%%%%%%%%%%%%%%%%%%%%%%%%%%%%%%%%%%%%%%%%%%%%%%%%%%%%%%

\newpage
\onecolumn
%\tableofcontents

\appendix
\section{Additional Analysis }
\subsection{Shift of Selected Backdoor Angle}
\label{sec:explain}

\begin{wrapfigure}{r}{0.4\textwidth}
\includegraphics[width=1\linewidth]{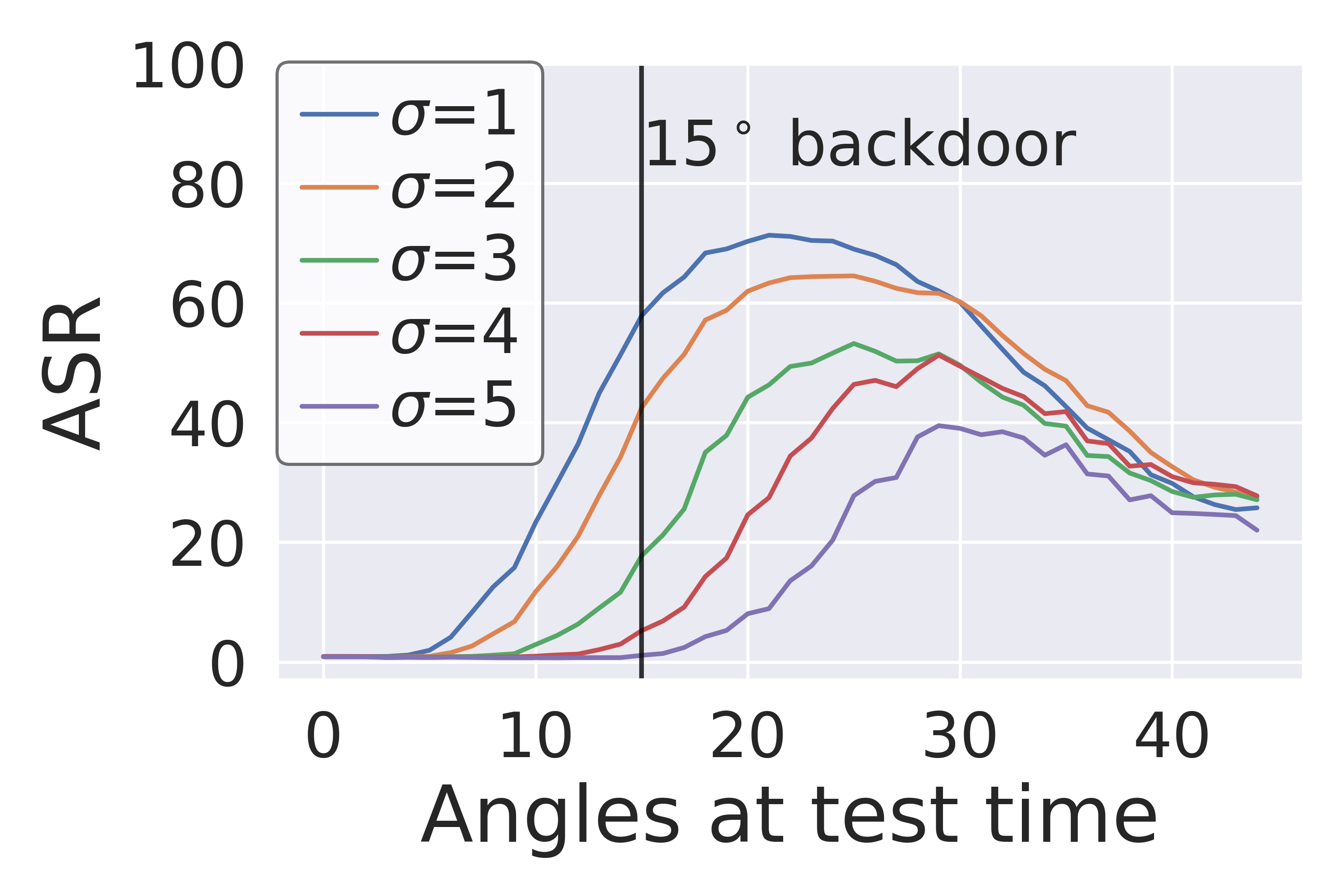}
\caption{ Explicitly increasing the variance of rotation angles in the original training data by randomly rotating each image with degree drawn from $\N(0, \sigma; [-180^{\circ}, 180^{\circ}])$.  }
\label{fig:angle-shift}
\end{wrapfigure}

As mentioned in Section \ref{sec:analysis}, an interesting phenomenon is that the most effective attack angle at the test time is usually slightly higher than the predefined backdoor angle. Here, we provide a simple explanation for this phenomenon. 
We formulate backdoor prediction as a hypothesis testing problem. We use $\N(\mu, \sigma^2; [a, b])$ to denote truncated Gaussian within the interval of $[a, b]$. 
We restrict the rotation degree to be within $[-180^{\circ}, 180^{\circ}]$.
%, so if the rotation degree becomes -200, it refers to 160. 
For a natural vision dataset, many images may already be rotated due to different camera viewpoints. 
We assume that the rotation degree of images in the original training distribution follows truncated Gaussian distribution
$\D \sim \N(0, \sigma^2; [-180^{\circ}, 180^{\circ}])$. Gaussian distribution is arguably the most reasonable assumption about rotation degrees in natural datasets due to the maximum entropy principle. Let $\beta$ denotes the backdoor angle degree inserted during the training time, and $\rho$ denotes the poisoning rate. Then for poisoned data, the distribution of rotation degree follows $\D_{\backdoor} \sim \N(\beta, \sigma^2; [\beta - 180^{\circ}, \beta + 180^{\circ}] )$. The overall rotation degree distribution after poisoning becomes a mixture of truncated Gaussian $(1-\rho) \D + \rho \D_{\backdoor}$.

We model the classification task as a hypothesis testing problem, where the neural network needs first to decide whether the inputs are drawn from $\D$ or $\D_{\backdoor}$, and then make the prediction accordingly. 
For an image with rotation angle of degree $x$, in order to minimize the cross entropy loss, the optimal classifier will predict clean label with probability $\frac{(1-\rho) \D(x)}{(1-\rho) \D(x)+\rho \D_{\backdoor}(x)}$, and backdoored target label with probability $\frac{\rho \D_{\backdoor}(x)}{(1-\rho) \D(x)+\rho \D(x)}$. 
Thus, the attack success rate for optimal classifier on rotation degree $x$ is upper bounded by $\frac{\rho \D_{\backdoor}(x)}{(1-\rho) \D(x) + \rho \D(x)}$. 
In the following theorem, we show that the maximum possible attack success rate monotonically increases with the attack angle at the test time due to the exponential decay property of the Gaussian distribution. 
\begin{theorem}
Given sufficient training data points, the attack success rate for the optimal classifier on backdoored image $x$ is maximized at $x=180^{\circ}$.  
\end{theorem}
However, due to the low density of Gaussian tails, there may not be enough data points with large rotation angles for training the optimal classifier. Therefore, the optimal backdoor angles at the test time are only moderately higher than the backdoor angle at training time. 
To further validate our theory, in Figure \ref{fig:angle-shift}, we increase the variance of the rotation degree of original training data by randomly rotating each data point with a degree drawn from $\N(0, \sigma; [-180^{\circ}, 180^{\circ}])$. In this case, there are more data points with a large rotation degree, which pushes the optimal backdoor angle to be higher. As shown in the figure, the optimal backdoor at the test time increases as $\sigma$ grows, which matches our explanation.

% Then the backdoored data distribution follows $\D_{\backdoor} \sim \N(\beta, \sigma^2; [\beta - 180, \beta + 180] )$, 
% and the overall rotation degree becomes a mixture of truncated Gaussian $(1-\rho) \D + \rho \D_{\backdoor}$. 
% The backdoor classification task can be modeled as a hypothesis testing problem, where the neural network needs to decide whether the inputs are drawn from $\D$ or $\D_{\backdoor}$. 
% For any image with a rotation degree $x$, in order to minimize the sum of false positive and false negative errors, the optimal classifier will predict clean label with probability $\frac{(1-\rho) \D(x)}{(1-\rho) \D(x)+\rho \D(x)}$, and backdoored target label with probability $\frac{\rho \D_{\backdoor}(x)}{(1-\rho) \D(x)+\rho \D(x)}$. 
% Thus, the attack success rate for optimal classifier on backdoored image $x$ is upper bounded by $\frac{(1-\rho) \D(x))}{(1-\rho) \D(x) + \rho \D(x)}$.
% \begin{theorem}
% Given sufficiently amount of training data points, the attack success rate for optimal classifier on backdoored image $x$ is maximized at $x=180$.  
% \end{theorem}

%, and we want to explain it through Bayes error. 

\subsection{ Effective Rotation Angle}
\label{sec:identify}

We further study the influence of backdoored rotation angles on the ASR. 
As \citet{Zhang2021HowTI} suggested, to ensure the existence of backdoored parameters, a backdoor pattern should follow: (1)  added on low-variance features, and (2) the strength of the backdoor pattern is enough. The first condition can be verified by our experiment in Section \ref{sec:data_aug_eff}, where rotation augmentation increases the variance of rotation, thereby mitigating the poisoning effect. 
We imply the second condition as that valid backdoor transformation should shift the samples far enough from the original inputs. Therefore, we adopt the loss of transformed data w.r.t. the naturally trained classifier to quantify the semantic difference. 

\begin{figure}[H]
\centering
\includegraphics[width=0.5\linewidth]{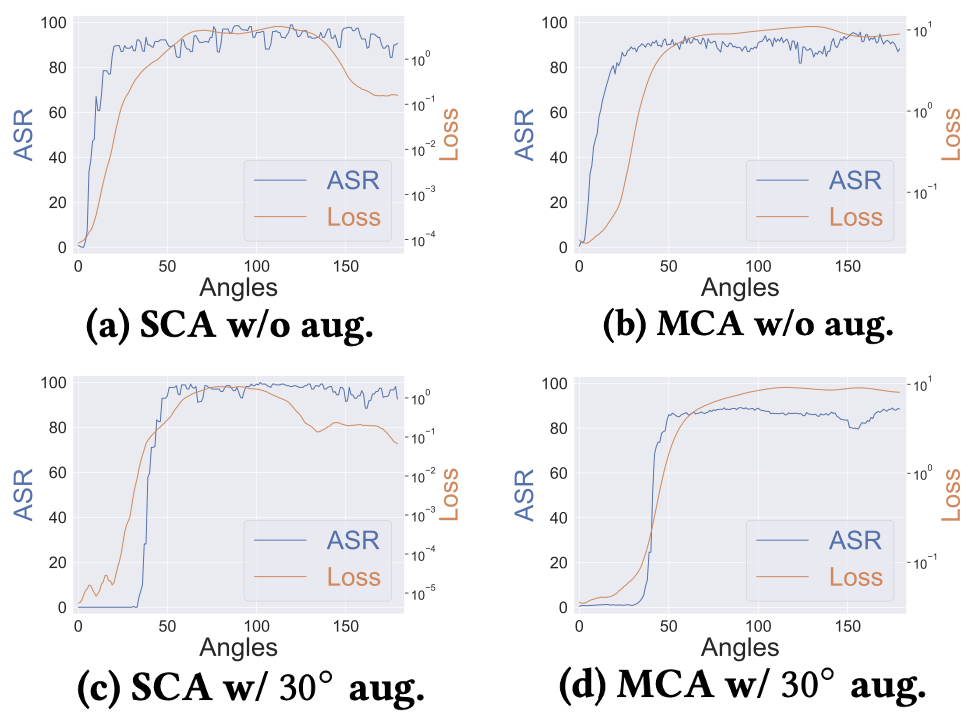}
\caption{Correlation between the loss of rotated images w.r.t. clean model and attack success rate. }
\label{fig:identify}
\end{figure}

\smallskip
\noindent \textbf{Attack success rate exhibits a steep rise as backdoored angle increases.} Figure \ref{fig:identify} visualizes the correlation between the loss and corresponding ASR from $0^\circ$ to $180^\circ$. We notice a sharp increase for ASR through a range of $\sim$20 degree; for example, in figure \ref{fig:identify}c, the attack success rate achieves almost 100\% by utilizing a 50-degree backdoor but turns to 0\% using 30 degrees one. We also observe that ASR generally exhibits a similar trend with the value of cross entropy loss in log scale, and we can identify that angles with larger than $10^{-1}$ loss value on the GTSRB dataset attain good attacking performance. With that, we bridge over the spatial robustness and rotation backdoor attacks.

\end{document}